\documentclass[10pt,twocolumn,letterpaper]{article}

\usepackage{cvpr}              

\usepackage{capt-of}
\usepackage{pgfplots}
\usepackage{multirow}
\usepackage{graphicx}
\usepackage{tikz}
\usepackage{calligra}
\newcommand{\editfont}{\ttfamily}  
\usepackage[outline]{contour}
\contourlength{0.2pt} 
\usepackage[normalem]{ulem} 
\usepackage{cuted}
\usepackage{multirow}
\usetikzlibrary{calc,positioning}
\usepackage{tikz}
\usepackage{pgfplots}
\usepgfplotslibrary{groupplots}
\pgfplotsset{compat=1.18}
\usepackage{algorithm}
\usepackage{algpseudocode}
\usepackage{bbm}
\newcommand{\myComment}[1]{\hfill\small\textcolor{gray}{$\triangleright$ \textit{#1}}}

\definecolor{cvprblue}{rgb}{0.21,0.49,0.74}
\usepackage[pagebackref,breaklinks,colorlinks,allcolors=cvprblue]{hyperref}

\usepackage[accsupp]{axessibility}  

\definecolor{lightgrayfill}{RGB}{245,245,245}
\definecolor{ourblue}{RGB}{0,133,255}
\definecolor{renoisered}{RGB}{230,0,35}
\definecolor{exactorange}{RGB}{255,140,0}
\definecolor{ddimgreen}{RGB}{0,155,0}

\usepackage{amsmath}
\def\paperID{46450} 
\def\confName{CVPR}
\def\confYear{2026}

\title{Prompt-Guided Image Editing with Masked Logit Nudging in Visual Autoregressive Models}

\author{
Amir El-Ghoussani$^{1}$ \quad 
Marc H{\"o}lle$^{1}$ \quad 
Gustavo Carneiro$^{2}$ \quad 
Vasileios Belagiannis$^{1}$ \\[6pt]
$^{1}$Friedrich-Alexander-Universität Erlangen-Nürnberg, Germany \\
$^{2}$University of Surrey, United Kingdom \\[2pt]
{\tt\small \{first.last\}@fau.de} \quad {\tt\small g.carneiro@surrey.ac.uk}
}

\begin{document}


\twocolumn[{%
  \renewcommand\twocolumn[1][]{#1}%
  \maketitle
  \vspace{-1.0em}
  \begin{center}
    \includegraphics[scale=.075]{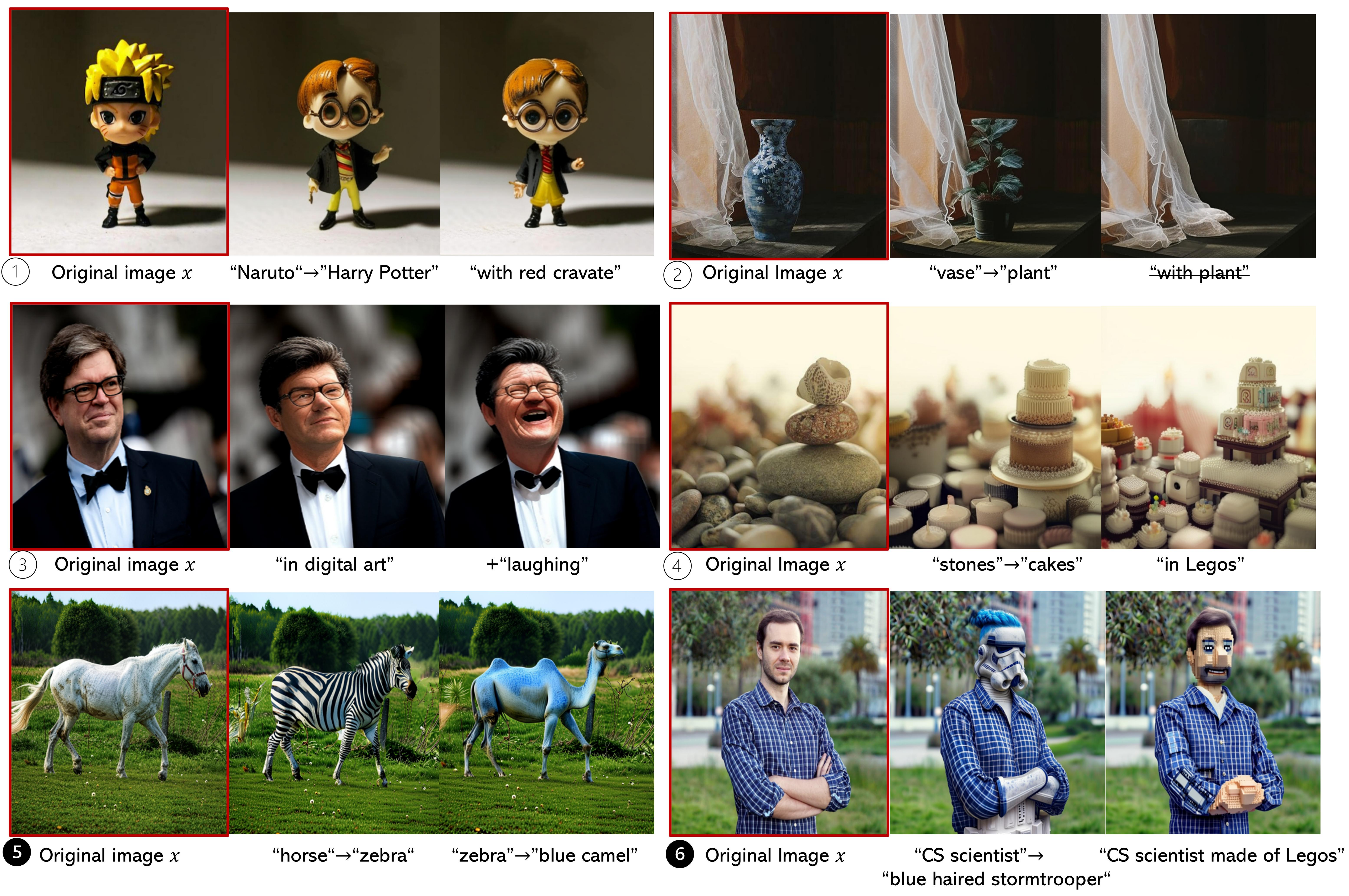}
    \vspace{0.6em}
    \captionsetup{type=figure*}
    \captionof{figure}{
      We present \textbf{Masked Logit Nudging} for image editing in Visual Autoregressive (VAR) models. 
      Given a source image and a target prompt, our method produces high-quality edited outputs while maintaining strong structural fidelity. 
      MLN effectively handles diverse editing types, including object removal (example~2), attribute addition (examples~1--2), attribute modification (examples~3 and~5), and style change (examples~4 and~6). 
      By softly correcting quantization errors during reconstruction, our approach achieves superior background preservation. 
      It enables real-time editing, processing $512{\times}512$ images in $\approx$ \textbf{0.82\,s} (examples~1--4, circled white numbers) and $1024{\times}1024$ images in $\approx$ \textbf{1.6\,s} (examples~5--6, circled black numbers), without any training or inversion. 
      Our framework is fully compatible with all VAR-based generative models.
    }
    \label{fig:teaser}
  \end{center}
  \vspace{1em}
}]

\begin{abstract}
    We address the problem of prompt-guided image editing in visual autoregressive models. Given a source image and a target text prompt, we aim to modify the source image according to the target prompt, while preserving all regions, which are unrelated to the requested edit. To this end, we present Masked Logit Nudging\footnote{\url{https://github.com/AmirMaEl/MLN}}, which uses the source image token maps to introduce a guidance step that aligns the model's predictions under the target prompt with these source token maps. Specifically, we convert the fixed source encodings into logits using the VAR encoding, nudging the model's predicted logits towards the targets along a semantic trajectory defined by the source–target prompts. Edits are applied only within spatial masks obtained through a dedicated masking scheme that leverages cross-attention differences between the source and edited prompts. Then, we introduce a refinement to correct quantization errors and improve reconstruction quality. Our approach achieves the best image editing performance on the PIE benchmark at 512px and 1024px resolutions. Beyond editing, our method delivers faithful reconstructions and outperforms previous methods on COCO at 512px and OpenImages at 1024px. Overall, our method outperforms VAR-related approaches and achieves comparable or even better performance than diffusion models, while being much faster.
\end{abstract}

\section{Introduction}
Recent advances in image generation have revolutionized visual synthesis and editing with paradigms such as diffusion models~\cite{rombach2022high, ho2020denoising} and rectified flows~\cite{lipman2022flow} achieving remarkable success. Their effectiveness in image-editing largely stems from inversion, i.e. recovering the noise that would have generated the image. However, this editing-by-inversion paradigm has well-documented shortcomings~\cite{brack2024ledits++, samuel2023lightning, huberman2024edit, deutch2024turboedit}. In practice, inversion errors propagate through the sampling process, producing unintended modifications and reducing fidelity to the source image~\cite{ju2023direct}. Even when the original noise is known exactly, such as when editing generated images, this approach frequently distorts local structures or global composition~\cite{huberman2024edit}. Attempts to mitigate these issues by refining inversion~\cite{ju2023direct, mokady2023null} or injecting intermediate representations, such as attention maps~\cite{hertz2023prompt}, can improve fidelity, but these solutions remain fragile, model-specific, and computationally costly.

The above limitations have motivated the exploration of alternative generative models, such as token-based autoregressive (AR) models, which were originally dominant in natural language processing~\cite{Touvron2023LLaMAOA}. Methods such as LlamaGen~\cite{sun2024autoregressive} improve image tokenization and transformer architecture design, reaching quality competitive with diffusion models while maintaining simple sampling. However, despite their architectural simplicity, plain autoregressive models remain slow because they generate images sequentially, token by token. This causes the cost of sampling to grow linearly with the number of image tokens and limits their scalability for high-resolution image generation and interactive editing. Visual autoregressive (VAR)~\cite{tian2024visual} models have recently gained popularity due to their ability to operate directly in latent space. These models predict entire token maps in a progressive coarse-to-fine manner, enabling higher throughput and spatial consistency.

Despite these advances, prompt-guided image editing within VAR approaches remains a challenge. Existing VAR-based editing methods rely on more restrictive or error-prone procedures. AREdit~\cite{wang2025training} depends on the BSQ tokenization scheme~\cite{zhaoimage} and therefore applies only to Infinity-style models~\cite{han2024infinity}, limiting its generality. VARIN relies on an argmax pseudo-inversion that, similar to diffusion and rectified-flow inversion, introduces errors that accumulate through the generative process.

In this work, we address these challenges by proposing an architecture-agnostic, inversion-free, and prompt-guided image editing approach for VAR models~\cite{tian2024visual}. Our goal is to modify the source image according to the target prompt while preserving all regions unrelated to the requested edit.

We propose \textit{Masked Logit Nudging}, a mechanism that guides the VAR model to perform prompt-driven image edits while maintaining fidelity to the source image.  
It makes use of the source token maps obtained from the original image and introduces a guidance step that aligns the model’s predictions under the target prompt with these source token maps.
By softly balancing~(\textit{``nudging''}) the model’s predicted outputs, conditioned on the target prompt, toward the source image structure and semantics, the proposed mechanism enables edits that follow the target prompt while preserving the overall visual consistency of the original image.
Furthermore, we extract the VAR Transformer's~\cite{tian2024visual} internal cross-attention maps by feeding the encoded source token maps into the VAR Transformer and conditioning it separately on the source and target prompts.  
These maps capture how different regions of the source image respond to words in each prompt.  
By comparing the attention responses between the source- and target-conditioned passes, we create a mask of attention changes. \textit{Masked Logit Nudging} is then applied only within this mask, ensuring that modifications are limited to regions that are semantically affected by the target prompt. To enhance reconstruction fidelity, we introduce a refinement that corrects quantization errors. These are small distortions or color shifts that occur when continuous image features are converted into discrete token maps during encoding and decoding. 
Combined, these components enable our mechanism to achieve accurate, prompt-aligned edits while preserving the overall layout and appearance of the original image.

Extensive evaluation shows that our mechanism achieves promising performance compared to VAR-related methods, and is comparable to or better than diffusion performance in image editing, while  being much faster. Our main contributions are summarized as follows:
\begin{itemize}
\item Masked Logit Nudging: An inversion-free, prompt-guided editing method that operates directly in logit space.
\item Cross-Attention–Driven Masking: A spatially aware masking scheme that leverages cross-attention differences between source and target prompts.
\item Quantization Refinement: A quantization-aware refinement for reducing reconstruction artifacts and improving visual fidelity during editing.
\item State-of-the-art image editing performance on the PIE benchmark at both 512px and 1024px. Beyond editing, our method also delivers faithful reconstructions, outperforming prior approaches on COCO at 512px and OpenImages at 1024px.

\end{itemize}

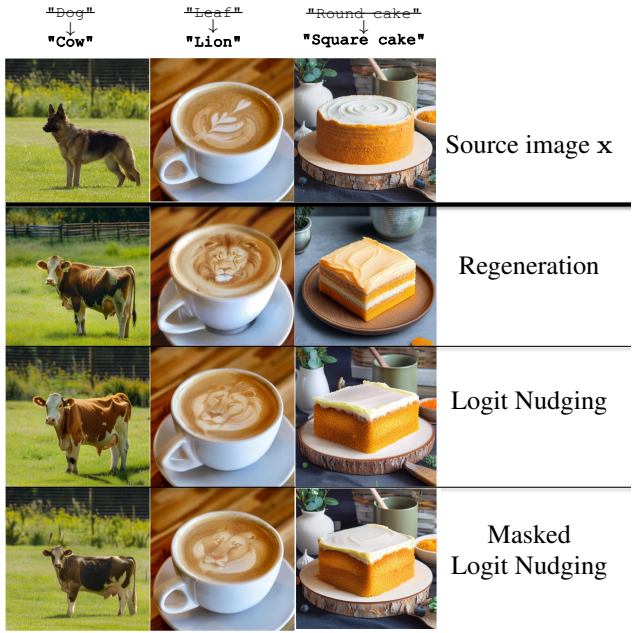
\begin{figure}[t]
\centering
\begin{tikzpicture}
  \node[inner sep=0] (image) {\includegraphics[width=\linewidth]{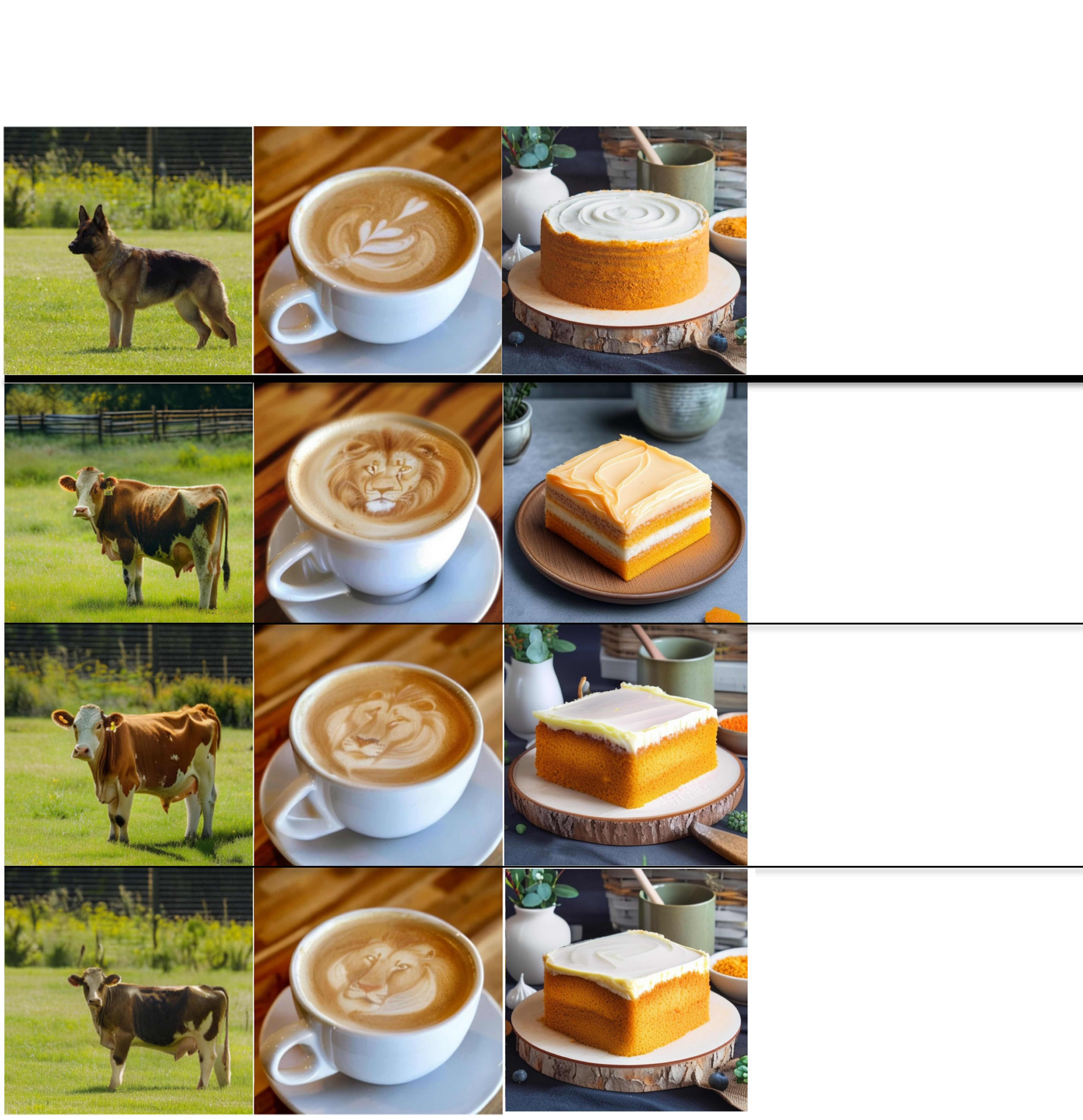}};

\begin{scope}[shift={(image.south west)}, x=1cm, y=1cm]

    \tikzset{
      colhead/.style={
        font=\fontsize{6pt}{5pt}\editfont,
      },
      rowlbl/.style={
        font=\normalsize
      }
    }

    \node[colhead, anchor=south, align=center] at (0.9, 7.7) {\sout{"Dog"} \\  $\downarrow$\\ \contour{black}{"Cow"}};
    \node[colhead, anchor=south, align=center] at (2.8, 7.7) {\sout{"Leaf"}\\ $\downarrow$ \\ \contour{black}{"Lion"}};
    \node[colhead, anchor=south, align=center] at (4.8,7.7) {\sout{"Round cake"}\\ $\downarrow$ \\ \contour{black}{"Square cake"}};

    \node[rowlbl, anchor=south] at (7, 6.2) {Source image $\mathbf{x}$};
    \node[rowlbl, anchor=south] at (7, 4.6) {Regeneration};
    \node[rowlbl, anchor=south] at (7,2.8) {Logit Nudging};
    \node[rowlbl, anchor=south, align=center] at (7, 0.6) {Masked \\Logit Nudging};

  \end{scope}
\end{tikzpicture}
\caption{\textbf{Qualitative comparison.} Edits generated by the proposed Regeneration, Logit Nudging, and Masked Logit Nudging, showing reduced unintended modifications in background regions compared to the source image.}
\label{fig:regeneration}
\end{figure}

\section{Related work}
We review prior work on text-guided image editing with focus on autoregressive modeling.

\paragraph{Text-guided Image Editing}
\label{sec:rw:text_guided}
Text-guided image editing allows to modify visual content through natural language prompts. Early diffusion-based approaches~\cite{hertz2023prompt,brack2024ledits++,brooks2023instructpix2pix} rely on inversion techniques~\cite{galimage,ju2023direct,mokady2023null} to recover structured noise from an input image and re-generate it according to a target prompt. While effective, these methods often suffer from inaccurate inversion and entangled text-image features, resulting in global, unintended changes. Subsequent works addressed these issues using attention control~\cite{hertz2023prompt,tumanyan2023plug}, rectified flows~\cite{routsemantic,wang2024taming}, or improved inversion solvers~\cite{brack2024ledits++}, but these remain computationally heavy due to iterative denoising and multi-step guidance. 
Moreover inversion-based editing applies the regular generative diffusion process and is therefore exposed to general reliability concerns of diffusion 
models, such as memorization issues~\cite{carlini2023extracting,asthana2026detecting}.

In contrast, our approach is fully inversion-free: we operate directly in logit space and achieve localized edits in a single forward pass, offering a high level of controllability at visual autoregressive efficiency.

\paragraph{Autoregressive Image Generation}
\label{sec:rw:ar_image_gen}
Autoregressive (AR) modeling, widely used in language modeling, has recently been extended to vision through token-based architectures such as VQGAN~\cite{esser2021taming} and VQVAE~\cite{van2017neural}. Subsequent large-scale models~\cite{yu2023language} demonstrated image quality comparable to diffusion models, but inference remains slow due to sequential token prediction. {VAR} models~\cite{tian2024visual} address this through a {next-scale prediction} scheme that generates images hierarchically from coarse to fine scales, greatly improving efficiency while maintaining visual fidelity. 
Building on this foundation, several recent works have advanced VAR-based architectures. {STAR}~\cite{ma2024star} introduced text-conditional next-scale generation for text-to-image synthesis, while {HART}~\cite{tang2024hart} combined visual autoregressive prediction with lightweight diffusion refinement for enhanced realism. {Infinity}~\cite{han2024infinity} proposed a bitwise quantization scheme that scales VAR to billion-parameter capacity, and {SWITTI}~\cite{voronov2024switti} further improved scalability by removing causal constraints, enabling high-resolution text-to-image generation at unprecedented speed. Beyond image synthesis, VAR priors have also been extended to dense prediction tasks such as monocular depth estimation~\cite{visapp26,gabdullin2025depthart,hornauer2025revisiting}.

Our work builds directly on VAR architectures but extends them toward controllable image editing, introducing spatially guided logit-level manipulation that previous generation-only VAR approaches do not support.

\paragraph{Image Editing with VAR Models}
\label{sec:rw:var_image_edit}
Despite the recent success of VARs in image synthesis, text-guided editing within these models remains largely unexplored. The first such method, {AREdit}~\cite{wang2025training}, introduced a training-free VAR editing pipeline that caches token distributions from the source image and applies adaptive probability masking to selectively re-sample edited regions. Although efficient, AREdit determines editable regions solely through probability differences and remains restricted to the VAR backbone Infinity~\cite{han2024infinity}. Concurrently, {VARIN}~\cite{dao2025discrete} proposed an inversion-based technique using a discrete {Location-Aware Argmax Inversion (LAI)} to reconstruct inverse noises for editing. While VARIN improves reconstruction fidelity, it relies on pseudo-inversion of non-invertible argmax operations, making it computationally expensive and unstable, and it lacks region-aware masking. 
{In contrast, our method introduces MLN—a direct, spatially controlled editing mechanism that requires no inversion or caching, applies edits only within cross-attention–derived masks, and preserves fidelity through quantization error refinement.}


\section{Method}
Given a source image $\mathbf{x}$ with a source prompt $t_s$ describing its content and a target prompt $t_t$ specifying the desired edit, our goal is to generate an edited image $\mathbf{y}$ that reflects the semantics of $t_t$ while preserving the structure of $\mathbf{x}$.
 To achieve this, we introduce an inversion-free, prompt-guided editing approach that operates directly in the latent token space of a pretrained text-to-image VAR model, enabling effective semantic manipulation without additional finetuning or model retraining.

We first provide background on VAR (Sec.~\ref{subsection:multi-token}) and introduce \textit{masked logit nudging} (Sec.~\ref{sec:logit}), a  guidance mechanism that steers the transformer’s predicted logits toward the semantics of the target prompt $t_t$. 
To further constrain modifications spatially, we propose a \textit{dedicated masking strategy} (Sec.~\ref{sec:m:guidance}) that identifies edit regions based on cross-attention differences between source and target prompts. 
Finally, we enhance VAR’s reconstruction fidelity via a \textit{quantization refinement} in the decoding process (Sec.~\ref{section:soft-projection}).

\subsection{Visual Autoregressive Modeling}
\label{subsection:multi-token}
VAR modeling~\cite{tian2024visual} formulates autoregressive image generation as \textit{next-scale prediction}, by utilizing a multi-scale visual tokenizer together with a decoder-only transformer. Specifically, an image is quantized into $K$ multi-scale token maps $R = (\mathbf{r}_1, \mathbf{r}_2, \ldots, \mathbf{r}_K)$, each with progressively higher spatial resolution $h_k \times w_k$. During generation, the transformer $\mathcal{T}(\cdot)$ predicts a whole token map $\mathbf{r}_k$ at scale $k$, conditioned on the sequence of lower-scale token maps $(\mathbf{r}_1, \mathbf{r}_2, \ldots, \mathbf{r}_{k-1})$. 

\paragraph{Encoding \& Decoding} Formally, during encoding, an encoder $\mathcal{E}$ transforms the image $\textbf{x}$ into a continuous feature representation $\textbf{f}=\mathcal{E}(\mathbf{x})\in \mathbb{R}^{h\times w\times d}$, where $h$, $w$, and $d$ denote the height, width, and channel dimension, respectively. Subsequently, the quantizer $\mathcal{Q}$ maps these continuous features into discrete token maps $R = (\mathbf{r}_1, \mathbf{r}_2, \ldots, \mathbf{r}_K)=\mathcal{Q}(\textbf{f})$. Intuitively, the first token map $\mathbf{r}_1$ captures a global representation with size $1 \times 1 \times d$, while the final map $\mathbf{r}_K \in \mathbb{R}^{h_K \times w_K \times d}$ corresponds to the full-resolution encoded representation, i.e., $h_K = h$ and $w_K = w$. During decoding, VAR progressively aggregates the sequence of token maps $\mathbf{r}_k$ to approximate the original feature representation $\mathbf{f}$ as:
\begin{equation}
\label{eq:featrepr}
    \hat{\mathbf{f}}=\sum_{k=1}^{K}\text{Up}(\text{Lookup}_{\mathbf{C}}(\textbf{r}_k),(h,w))=\sum_{k=1}^{K}\text{Up}(\textbf{f}_k, (h,w)), 
\end{equation}
where $\text{Lookup}_C(\textbf{r}_k)$ retrieves the continuous vector representations $\hat{\textbf{f}}_k$ from the shared codebook $C = \{c_1, \dots, c_V\}$ with codebook size  $V$ and $\text{Up}(\cdot,(h\times w)$ upsamples the vector representations to resolution $h\times w$. Finally the decoder $\mathcal{D}$ processes the combined feature representation $\hat{\mathbf{f}}$ to produce the final decoded output, such that $\hat{\mathbf{x}} = \mathcal{D}(\hat{\mathbf{f}})$. 
We adopt SWITTI~\cite{voronov2024switti} as our main VAR model. In SWITTI, the transformer autoregressively predicts the likelihood of scale $k$'s token map $\hat{\mathbf{r}}_k$ based on the previous token map $\hat{\textbf{r}}_{k-1}$ and the CLIP~\cite{radford2021learning} text embeddings $\psi(t)$ of a text prompt $t$ according to:
\begin{equation}
p\left(\mathbf{r}_1, \mathbf{r}_2, ..., \mathbf{r}_k|\psi(t)\right) = \prod_{k=1}^K p(\mathbf{r}_k | \mathbf{r}_{k-1},\psi(t)).
\end{equation}
\paragraph{Sampling} 
At each scale $k$, the transformer $\mathcal{T}(\cdot)$ produces a logit tensor $\hat{\mathbf{z}}_k \in \mathbb{R}^{(h_k \times w_k) \times V}$, where each element corresponds to a categorical distribution over the $V$ codebook entries in $\mathbf{C}$.
We apply a softmax operation along the codebook dimension to obtain normalized token probabilities, $\mathrm{softmax}(\hat{\mathbf{z}}_k)$, and sample the final token indices $\hat{\mathbf{r}}_k$ using standard autoregressive sampling strategies, including top-$k$, nucleus sampling, or Gumbel-softmax~\cite{topp, jang2017categorical}.
\begin{figure*}[t]
\centering
\begin{tikzpicture}
  \node[inner sep=0] (image) {
    \includegraphics[scale=.066]{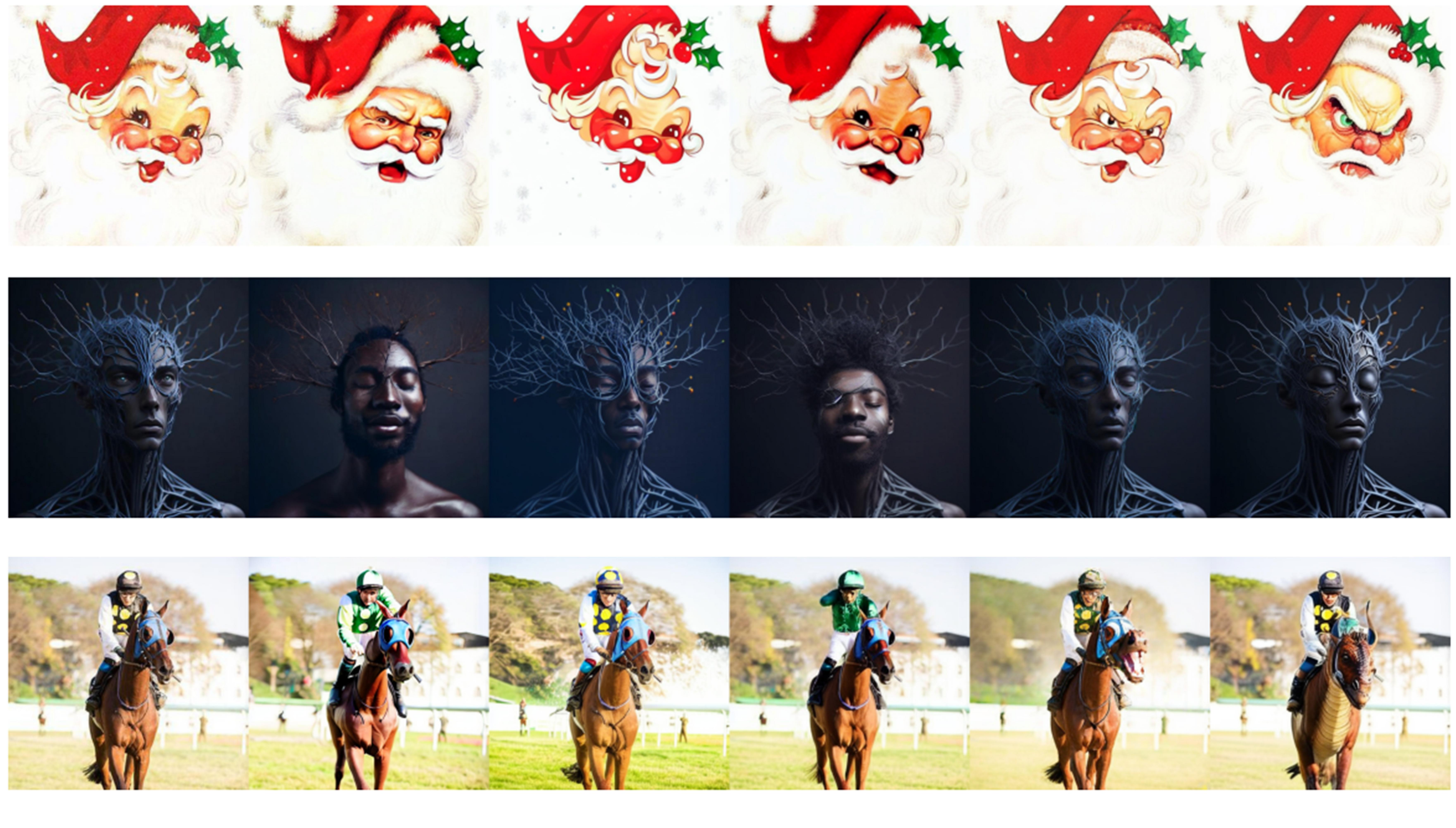}
  };

  \begin{scope}[shift={(image.south west)}, x=1cm, y=1cm]

    \usetikzlibrary{positioning}

    \tikzset{
      colhead/.style={
        font=\fontsize{7pt}{5pt}\editfont,
      },
      rowlbl/.style={
        font=\normalsize,
      }
    }


    \node[rowlbl, anchor=south] (src) at (1.3, 9.3) {Source image $\mathbf{x}$};
    \node[rowlbl, right=.3cm of src] (edf) {EditFriendly~\cite{huberman2024edit}};
    \node[rowlbl, right=1cm of edf] (pnp) {PnP~\cite{ju2023direct}};
    \node[rowlbl, right=.8cm of pnp] (ledits) {Ledits++~\cite{brack2024ledits++}};
    \node[rowlbl, right=.3cm of ledits, align=center] (turbo)
      {TurboEdit~\cite{deutch2024turboedit}};
    \node[rowlbl, right=1.0cm of turbo, align=center] (ours)
      {Ours};

    \node[colhead,  anchor=north,above=0cm of src] at (image.south)
      {"A jockey rides a [\sout{horse} $\rightarrow$\contour{black}{dragon}]".};
    \node[colhead,  anchor=north,above=3cm of src] at (image.south)
      {"A man with a tree head and branches coming out his ear [$\rightarrow$\contour{black}{with his eyes shut}]".};
    \node[colhead,  anchor=north,above=6.15cm of src] at (image.south)
      {"A christmas illustration of santa's [\sout{laughing} $\rightarrow$\contour{black}{angry}] face".};
  \end{scope}
\end{tikzpicture}
\caption{\textbf{Qualitative results.} Editing results of EditFriendly~\cite{huberman2024edit}, PnP~\cite{ju2023direct}, Ledits++~\cite{brack2024ledits++}, TurboEdit~\cite{deutch2024turboedit}, and our proposed Masked Logit Nudging (Ours). Masked Logit Nudging produces high-fidelity edits while minimizing unintended background modifications, such as blurring or structural changes.}

\label{fig:qualrec512}
\end{figure*}
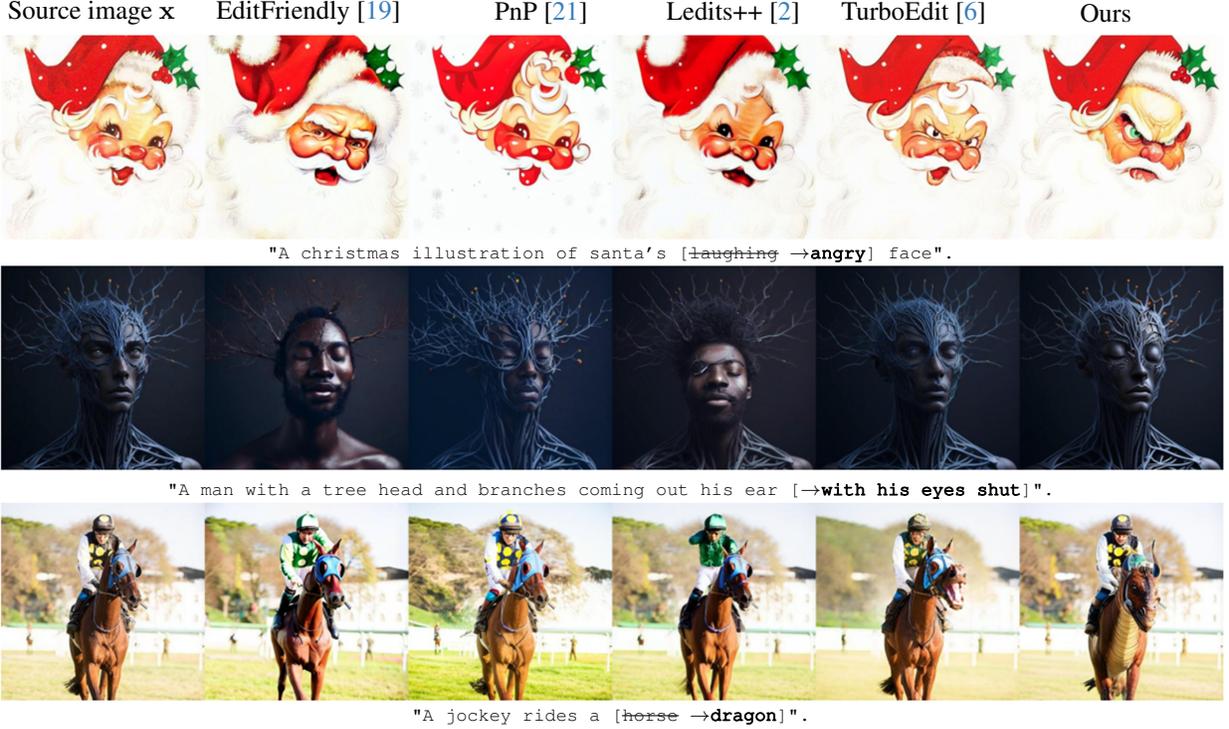

\subsection{Masked Logit Nudging}
\label{sec:logit} 
Prompt-guided image editing is performed by first computing the multi-scale token maps $(\mathbf{r}_1, \ldots, \mathbf{r}_K)$ of the source image $\mathbf{x}$.
To enable controlled edits, we fix the first $s$ token maps $(\mathbf{r}_1, \ldots, \mathbf{r}_s)$ from the source image and autoregressively generate the remaining maps $(\hat{\mathbf{r}}_{s+1}, \ldots, \hat{\mathbf{r}}_{K})$ conditioned on the target prompt $t_t$. Formally, for scales $k > s$ the model samples:
\begin{equation}
\label{eq:regeneration}
\hat{\mathbf{r}}_k \sim p(\hat{\mathbf{r}}_k \mid \hat{\mathbf{r}}_{<k},, \psi(t_t)).
\end{equation}
Using Eq.~\ref{eq:featrepr}, the modified sequence of token maps
$(\mathbf{r}_1, \ldots, \mathbf{r}_s, \hat{\mathbf{r}}_{s+1}, \ldots, \hat{\mathbf{r}}_K)$
is then decoded into a continuous feature representation $\hat{\mathbf{f}}$, and the final edited image is obtained as $\mathbf{y} = \mathcal{D}(\hat{\mathbf{f}})$.
We refer to this process as regeneration.
As illustrated in Fig.~\ref{fig:regeneration}, plain regeneration provides no spatial control: the influence of the target prompt is not confined to specific regions, leading to undesired global changes and excessive structural modifications.

\paragraph{Logit Nudging} To enhance spatial controllability while maintaining prompt alignment, we draw inspiration from {classifier-free guidance} (CFG)~\cite{ho2022classifier}, a mechanism commonly used in diffusion models that steers the denoising trajectory by interpolating between unconditional $\mathbf{\hat{y}}_u$ and conditional predictions $\mathbf{\hat{y}}_c$. CFG amplifies the influence of the conditioned predictions by the guidance $\alpha$ according to:
\begin{equation}
\label{eq:cfg}
    \mathbf{\hat{y}}=\mathbf{\hat{y}}_u + \alpha \underbrace{(\mathbf{\hat{y}}_c - \mathbf{\hat{y}}_u)}_{\text{guidance direction}}
\end{equation}
We adopt this principle for prompt-guided editing in visual autoregressive modeling by interpolating between the model’s current prediction under the target prompt $t_t$, i.e., $(\hat{\mathbf{r}}_{s+1}, \ldots, \hat{\mathbf{r}}_K)$, and the source tokens $(\mathbf{r}_{s+1}, \ldots, \mathbf{r}_K)$ obtained from the source image $\mathbf{x}$. This procedure effectively pulls the predicted logits at higher scales ($k>s$) toward the source structure, while still maintaining alignment with the target prompt semantics.

Formally, let $\hat{\mathbf{z}}_k \in \mathbb{R}^{h_k \times w_k \times V}$ denote the predicted logits at scale $k$ conditioned on the target prompt $t_t$. In standard autoregressive generation, a discrete token index is typically selected from $\hat{\mathbf{z}}_k$ (e.g., via $\mathrm{argmax}$ for greedy decoding), collapsing the prediction into a one-hot representation and discarding the underlying probability structure.
Instead, we retain the full categorical distribution
\begin{equation}
    p(\textbf{r}_k \mid \textbf{r}_{<k},\, \Psi(t_t)) = \mathrm{softmax}(\mathbf{\hat{z}}_k )
\end{equation}
as a \textit{soft token representation}. This preserves the entire probability structure and enables continuous interpolation between the target prompt-guided prediction $\mathrm{softmax}(\hat{\mathbf{z}}_k)$ and the one-hot encoded source tokens $\mathbf{e}(\mathbf{r}_k)$.

Accordingly, we define \textit{logit nudging} at scale $k$ with nudging strength $\alpha_k$ as:
\begin{equation}
\label{eq:logit-nudging}
\tilde{\mathbf{z}}_k = \hat{\mathbf{z}}_k +\alpha_k \underbrace{(\mathbf{e}(\textbf{r}_k)-\mathrm{softmax}(\hat{\mathbf{z}}_k))}_{\text{nudging direction}}.
\end{equation}

Here, both $\hat{\mathbf{z}}_k$ and the output logits $\tilde{\mathbf{z}}_k$ reside in logit space, while the nudging direction is defined in probability space as the difference between the one-hot source token distribution $\mathbf{e}(\mathbf{r}_k)$ and the model’s soft prediction $\mathrm{softmax}(\hat{\mathbf{z}}_k)$. Unlike classical CFG (Eq.~\ref{eq:cfg}), which operates in data space and interpolates between unconditional and conditional predictions, our formulation performs guidance in probability space using the source tokens themselves as the conditional signal.\footnote{In our formulation, the role of the conditional prediction $\hat{\mathbf{y}}_c$ in classical CFG (Eq.~\ref{eq:cfg}) is replaced by the one-hot encoded source tokens $\mathbf{e}(\mathbf{r}_k)$ in Eq.~\ref{eq:logit-nudging}.}
\paragraph{Nudging Schedule}
In practice, we control the influence of logit-nudging using the nudging strength $\alpha_k$, which is applied at each scale $k$ following a predefined decay schedule (see supplementary material \ref{decay_schedule}).
Intuitively, $\alpha_k$ determines how strongly the logits are steered toward the source token~$e(\mathbf{r}_k)$ in Eq.~\ref{eq:logit-nudging} at each scale. To balance structural preservation and edit flexibility, we employ a decreasing schedule across scales: large $\alpha_k$ values are used at the early, coarse stages to maintain the overall spatial layout of the source image $\mathbf{x}$. Smaller values are applied at the finer, high-resolution stages to allow more localized modifications.

\subsection{Cross-Attention-Driven Masking}
\label{sec:m:guidance}
While plain logit nudging (Eq.~\ref{eq:logit-nudging}) improves fidelity to the source structure, it can still cause unintended modifications in background regions~(see Fig.~\ref{fig:regeneration}, logit nudging). To address this, we introduce a \textit{spatially restricted guidance mechanism} using a binary edit mask $\textbf{M}_k$, which localizes the influence of the target prompt $t_t$. The mask is derived from cross-attention differences between the source and target prompts $t_s$ and $t_t$, ensuring that edits are applied only to semantically relevant regions.

To compute $\mathbf{M}$, we extract cross-attention maps from two separate regeneration passes~(Eq.~\ref{eq:regeneration}): one conditioned on the source prompt $t_s$ and another on the target prompt $t_t$.  
During each pass, we fix all lower-scale tokens up to an empirically selected scale $s$ and autoregressively reconstruct the remaining higher-scale tokens while recording all cross-attention activations throughout the transformer decoder blocks. We provide later an ablation selecting $s$.

This yields a hierarchy of attention maps $\mathcal{\mathbf{A}}^{s}_{k}$ and $\mathcal{\mathbf{A}}^{t}_{k}$ for the source and target prompts, respectively, across scales $k$ and number of transformer heads $T$:
$\mathcal{\mathbf{A}}^{s}_k,\, \mathcal{\mathbf{A}}^{t}_{k} \in \mathbb{R}^{h_k\times w_k\times T}$.
We normalize each cross-attention map to the range $[0,1]$ and compute the absolute difference between the source and target attentions to identify spatial regions of semantic change. For each scale $k$, we aggregate the differences across transformer heads $T$ to obtain a per-token difference map:
\begin{equation}
\mathcal{\mathbf{D}}_k = \frac{1}{T}\,\big\| \mathcal{\mathbf{A}}_k^s - \mathcal{\mathbf{A}}_k^t \big\|_{1}\;\in\;\mathbb{R}^{h_k\times w_k}.
\end{equation}
Each $\mathbf{D}_k$ is subsequently normalized and thresholded to produce a binary edit mask $\mathbf{M}_k$. Specifically, we retain the top-$q$ percentile of high-difference pixels (e.g., $q=80$) to form:
\begin{equation}
\mathbf{M}_k = \mathbf{1}\!\left[\, \mathbf{D}_k\;>\;\mathrm{Quantile}\!\left(\mathbf{D}_k,\,q\%\right) \right].
\end{equation}
Intuitively, this procedure effectively identifies regions where the cross-attention response varies most between $t_s$ and $t_t$, highlighting areas likely to require semantic modification.

For our final masked logit nudging, we define two complementary binary masks: 
$\textbf{M}_k\in\{0,1\}^{h_k \times w_k}$ and $\overline{\textbf{M}}_k=\mathbf{1}-\textbf{M}_k$,
where $\textbf{M}_k$ indicates edit regions and $\overline{\mathbf{M}}_k$ marks regions to be preserved. Importantly we linearly interpolate the masks to the individual scale dimensions $h_k\times w_k$.
We then apply logit nudging within the edit region and strong preservation elsewhere to obtain the output logit $\tilde{\mathbf{z}}_k$:
\begin{equation}
\tilde{\mathbf{z}}_k \;=\; \mathbf{\hat{z}}_k \;+\; \Big( \beta\,\overline{\textbf{M}}_k+ \alpha_k\,\textbf{M}_k \Big)\odot 
\underbrace{\big(\mathbf{e}(\mathbf{r}_k)-\mathrm{softmax}(\hat{\mathbf{z}}_k)\big)}_{\text{nudging direction}}.
\end{equation}
We keep $\beta$ fixed across all scales $k$ and initialize it with the maximum $\alpha_k$ value to maintain consistent guidance toward the source token distribution. 
As shown in Fig.~\ref{fig:regeneration}, this yields localized edits without overwriting unedited regions.
\subsection{Quantization Refinement}
\label{section:soft-projection}
In image editing, all modifications are performed in latent space rather than pixel space, making accurate latent reconstructions crucial. As described in Sec.~\ref{subsection:multi-token}, images are encoded into a quantized latent representation that maps continuous features onto a discrete codebook. We observe that reconstructions of encoded images accumulate {quantization errors} across scales (Sec.~\ref{quant_errors}).

Quantization errors arise because the encoder must approximate a continuous feature $\mathbf{f}$ using the nearest codebook vector from a finite set of learned embeddings. This discretization introduces a reconstruction gap between the continuous feature and its quantized counterpart, leading to small but perceptible deviations that propagate across scales during decoding. We track these quantization errors by accumulating the residual discrepancies at each codebook lookup during encoding as:
\begin{equation}
    \mathbf{f}_{\text{rest}} = \sum_{k=1}^K (\mathbf{f} - \mathbf{f}_k),
\end{equation}
where $\mathbf{f}_k$ is the quantized feature at scale $k$.  
A naïve way to improve reconstruction would be to add $\mathbf{f}_{\text{rest}}$ back into the final feature map. However, this introduces strong artifacts because $\mathbf{f}$ lies on the manifold of learned codebook embeddings—vectors the decoder is trained to interpret—whereas the raw residual $\mathbf{f}_{\text{rest}}$ lies off this manifold and contains feature directions the decoder cannot decode properly.


Instead of directly adding the residual $\mathbf{f}_{\text{rest}}$, we iteratively project it back onto the codebook embedding space before combining it with the final feature representation $\mathbf{f}$. 
Each iteration $j$ projects the current residual into a form the decoder can interpret, updates the reconstruction, and then recomputes the remaining residual. Repeating this “project–update’’ cycle gradually removes off-manifold components while retaining useful corrections.To preserve the intended edits, this reprojection is applied only outside the edit mask $\mathbf{m}$, i.e., in regions that are not modified by the target prompt. 
Equations are further explained in the supplementary material~\ref{quant_refinement}.
\begin{figure*}[ht]
    \centering
    \includegraphics[width=\textwidth]{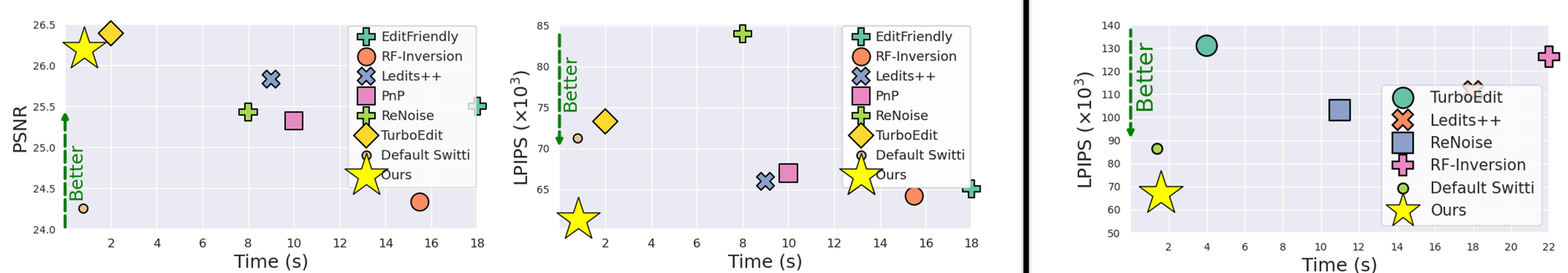} 
    \caption{
    \textbf{Reconstruction performance across resolutions.}
    \emph{Left:} PSNR and LPIPS averaged over 5{,}000 COCO validation images at $512\times512$ resolution, showing the trade-off between reconstruction fidelity and wall-clock time. 
    \emph{Right:} LPIPS averaged over 1{,}000 OpenImages samples at $1024\times1024$ resolution. 
    Across both benchmarks, our method achieves the best balance between image fidelity (higher PSNR, lower LPIPS) and computational efficiency.
    }
    \label{fig:reconstruction}
\end{figure*}
 
\begin{table*}[h]
\centering
\scriptsize
\caption{
\textbf{Quantitative evaluation on the PIE-Benchmark~\cite{ju2023direct} at 512$\times$512 resolution.}
We report background preservation (PSNR, LPIPS, MSE, SSIM), text–image alignment (CLIP similarity), and efficiency (inverse/forward time in seconds).
The \textit{Backbone} column specifies the underlying model family used by each method. \textbf{Bold} indicates the best performance.
}
\label{tab:piebench}
\resizebox{\textwidth}{!}{
\begin{tabular}{lcccccccccc}
\toprule
 & & \multicolumn{4}{c}{\textbf{Background Preservation}} 
 & \multicolumn{2}{c}{\textbf{CLIP Similarity}} 
 & \multicolumn{2}{c}{\textbf{Efficiency}} \\
\cmidrule(lr){3-6} \cmidrule(lr){7-8} \cmidrule(lr){9-10}
\textbf{Method} 
& \textbf{Backbone} 
& \textbf{PSNR$\uparrow$} 
& \textbf{LPIPS$\downarrow$ ($\times10^3$)} 
& \textbf{MSE$\downarrow$ ($\times10^4$)} 
& \textbf{SSIM$\uparrow$ ($\times10^2$)} 
& \textbf{Whole$\uparrow$} 
& \textbf{Edited$\uparrow$} 
& \textbf{Inverse (s)$\downarrow$} 
& \textbf{Forward (s)$\downarrow$} \\
\midrule
DDIM~\cite{ho2020denoising}             & SD1.4~\cite{rombach2022high}        & 17.88 & 197.96 & 196.66 & 76.86 & 23.62 & 21.20 & 24.9 & 26.2 \\
ReNoise~\cite{garibi2024renoise}        & SDXL~\cite{podellsdxl}        & 25.21 & 43.25  & 34.23  & 75.30 & 23.68 & 21.19 & 4.41 & 4.38 \\
LEDITS++~\cite{brack2024ledits++}       & SD1.5~\cite{rombach2022high}       & 26.54 & 45.88  & 39.45  & 69.18 & 24.97 & 22.23 & 5.11 & 4.59 \\
EditFriendly~\cite{huberman2024edit}    & SD1.4~\cite{rombach2022high}       & 26.33 & 90.08  & 91.48  & 85.17 & 23.97 & 22.03 & 15.00 & 20.00 \\
DirectInv~\cite{ju2023direct}           & SD1.4~\cite{rombach2022high}       & 27.12 & 62.55  & 58.81  & 86.77 & 25.33 & 22.40 & 20.12 & 15.13 \\
TurboEdit~\cite{deutch2024turboedit}    & SDXL-Turbo~\cite{sauer2024adversarial}  & 29.52 & 44.74  & 26.08  & \textbf{91.59} & 25.05 & 22.34 & 0.981 & 0.969 \\
\midrule
\multicolumn{10}{l}{\textbf{VAR / discrete-based:}} \\
\midrule
AREdit~\cite{wang2025training}         & Infinity~\cite{han2024infinity}    & 24.19 & 87.70  & --     & 83.40 & 25.60 & 22.70 & 0.50 & 0.50 \\
VARIN~\cite{dao2025discrete}            & HART~\cite{tang2024hart}& 26.54 & 54.04  & 38.33 & 85.39 & 25.40 & 21.30 & 4.00 & 2.00 \\
DICE~\cite{he2024dice}                  & Paella~\cite{rampas2022novel}      & 27.90 & 52.90  & 43.70 & 89.79 & 23.90 & 21.20 & --   & -- \\
\textbf{Ours}                           & SWITTI~\cite{voronov2024switti}      & \textbf{29.70} & \textbf{36.50} & \textbf{23.30} & 86.80 & \textbf{26.15} & \textbf{22.72} & \textbf{0.41} & \textbf{0.41} \\
\bottomrule
\end{tabular}
}
\end{table*}

\section{Experiments}


We first evaluate image editing performance (Sec.~\ref{sec:exp:editing}), followed by an analysis of reconstruction quality (Sec.~\ref{subsec:datasets_reconstruct}). We then assess the generality of our approach on an alternative VAR backbone (Sec.~\ref{sec:ablation_efficiency}). Due to space limitations, further ablations, hyperparameter studies, precision analysis, and qualitative results (including failure cases) are provided in the supplementary material~(Sec \ref{hyperparameter}).

\paragraph{Implementation Details}
\label{sec:implementation_detail}
We employ the pretrained {SWITTI}~\cite{voronov2024switti} text-to-image VAR model as our frozen backbone for our experiments. For 512px resolution we fix the lower-resolution tokens up to scale $s=6$ ($K=10$), and for 1024px resolution up to $s=8$ ($K=14$), following the scale hierarchy of SWITTI. Additionally for image editing, we disable quantization refinement (Sec.~\ref{section:soft-projection}) for style-based edits, since the refinement step feeds corrections back into the edited image, which can unintentionally alter colors and textures and thereby distort the intended style edit. Finally, the reconstruction experiments run with $\mathbf{M}_k=\mathbf{0}$.

\subsection{Image Editing}
\label{sec:exp:editing}

\paragraph{Datasets}
We evaluate image editing performance using the {PIE-Benchmark}~\cite{ju2023direct}, a standardized dataset designed to assess prompt-based image editing. It contains 700 images spanning 10 diverse editing scenarios such as object replacement, attribute modification, style transfer, and background alteration. Each sample is paired with source and target prompts and includes ground-truth editing masks for quantitative comparison. We conduct experiments at $512\times512$ resolution using the original PIE images and prompts. For $1024\times1024$ resolution, we construct an upscaled variant of the benchmark by applying the diffusion-based super-resolution model {InvSR}~\cite{yue2025arbitrary} to all images. The corresponding source and target prompts remain identical to the original setup. We prefer learned upscaling over simple interpolation because it restores plausible high-frequency details rather than merely enlarging pixels~(more details on the adapted benchmark in supplementary material Sec.~\ref{sr_benchmark}).

\begin{table}[t]
\centering
\setlength{\tabcolsep}{3.5pt}
\renewcommand{\arraystretch}{1.05}
\caption{\textbf{Quantitative evaluation on the (upscaled) PIE-Benchmark~\cite{ju2023direct} at 1024$\times$1024 resolution.}
We report background preservation (PSNR, LPIPS), text–image alignment (CLIP similarity), and wall-clock time. Best values are \textbf{bold}.}
\label{tab:pie_1024_small}
\scriptsize
\begin{tabular}{lcccc}
\toprule
\multicolumn{3}{c}{\textbf{Background Preservation}} &
\multicolumn{1}{c}{\textbf{CLIP Sim. $\uparrow$}} &
\multirow{2}{*}{\textbf{Wall (s) $\downarrow$}} \\
\cmidrule(lr){1-3}\cmidrule(lr){4-4}
\textbf{Method} & \textbf{PSNR$\uparrow$} & \textbf{LPIPS$\downarrow$} & \textbf{Whole / Edited} &  \\
\midrule
PnP~\cite{ju2023direct}                  & 19.59 & 117.96 & 23.62 / 21.20 & 17 \\
LEDITS++~\cite{brack2024ledits++}        & 23.32 & 82.65  & 23.97 / 21.03 & 18.4 \\
ReNoise~\cite{garibi2024renoise}         & 22.14 & 102.30 & 24.34 / 21.15 & 13.2 \\
TurboEdit~\cite{deutch2024turboedit}     & \textbf{27.62} & 34.33 & 25.23 / 23.56 & 4.1 \\
RF-Inversion~\cite{routsemantic}         & 21.22 & 67.43 & 24.41 / 22.11 & 22.3 \\
\textbf{Ours}                            & 26.70 & \textbf{31.50} & \textbf{26.81} / \textbf{23.67} & \textbf{1.6} \\
\bottomrule
\end{tabular}
\end{table}

\paragraph{Evaluation Metrics}
\label{sec:pie_metrics}
We evaluate the editing performance based on the protocol of the {PIE-benchmark}~\cite{ju2023direct}, which assesses reconstruction fidelity, perceptual similarity, and text alignment. For fidelity, we report the {Peak Signal-to-Noise Ratio (PSNR)} and the {Learned Perceptual Image Patch Similarity (LPIPS)}~\cite{zhang2018perceptual}, measuring pixel-level accuracy and perceptual consistency with the ground-truth image, respectively. To measure semantic alignment with the target prompt, we use the {CLIP similarity}~\cite{radford2021learning} between the edited image and the textual description, reported for both the whole image and the edited region. 
Finally, we report the {wall-clock time per edit} to quantify practical efficiency. 

\paragraph{Quantitative results}
Tab.~\ref{tab:piebench} and~\ref{tab:pie_1024_small} summarize performance on the 
PIE-Benchmark~\cite{ju2023direct} at $1024\times 1024$ and $512\times512$ resolution.  
At {1024px}~(Tab.~\ref{tab:pie_1024_small}) our method achieves the best perceptual quality (lowest LPIPS), the highest text-image alignment~(largest CLIP scores), and the fastest runtime (1.6\,s), outperforming diffusion and flow methods by an order of magnitude. At {512px}~\ref{tab:piebench}, our approach attains the strongest background preservation (best PSNR, LPIPS, and MSE) and the highest CLIP similarity, while also being the fastest method 
(0.82\,s).

\subsection{Reconstruction Quality}
We also evaluate MLN in the zero-edit setting, where the source and target prompts coincide. 
This tests whether the method can reproduce the input image without introducing unintended changes. In the following we provide more details on the conducted experiments to assess the reconstruction capability of our approach.


\paragraph{Datasets}
\label{subsec:datasets_reconstruct}
We evaluate reconstruction capability of our model at both 512\,px and 1024\,px. 
For the 512\,px setting, we use the COCO validation split~\cite{lin2014microsoft}, which 
contains 5{,}000 images, using the provided captions as both source and target prompts. For high-resolution evaluation, we introduce an {upscaled evaluation protocol} based 
on OpenImages~\cite{kuznetsova2020open}. Since OpenImages does not providecaptions or square image crops, we construct a new evaluation subset by filtering the training split for images larger than $1024\times1024$ with near-square aspect ratios, 
resizing them to $1024\times1024$, and using them for reconstruction benchmarking. 
Because captions are absent, we generate source/target descriptions using GPT-4V~\cite{zhang2023gpt} (see Supplementary Sec.~\ref{recaptioning}). 



\paragraph{Evaluation Metrics}
To quantitatively assess reconstruction quality, we evaluate the similarity between the original source image $x$ and the reconstructed image $\hat{x}$ using PSNR and LPIPS. In addition to these quantitative measures, we record the wall-clock time required to perform a complete reconstruction cycle. All reported values represent averages computed over the entire validation dataset.

\paragraph{Quantitative results.}
Fig.~\ref{fig:reconstruction} shows reconstruction quality versus runtime at 512px and 1024px. At 512px (fig.~\ref{fig:reconstruction}, left), we report PSNR and LPIPS averaged over $5000$ COCO images. Our method achieves low error and high PSNR while being among the fastest methods, yielding the best overall fidelity. At 1024px (Fig.~\ref{fig:reconstruction}, right), LPIPS averaged over $1000$ OpenImages samples again shows our method achieving the lowest perceptual error with the shortest runtime, outperforming all related methods.

\subsection{Ablation Studies}
\label{sec:ablation_efficiency}

\paragraph{Applicability to Other VAR Models}
To test how well \textit{Masked Logit Nudging} generalizes beyond our main backbone, we apply it to the {Infinity} model~\cite{han2024infinity} without any retraining or architectural changes. As shown in Table~\ref{tab:ablation_backbones}, MLN produces consistent behaviour across both backbones, confirming that its logit-space formulation transfers reliably to different VAR architectures. Due to space limitations, further ablations  are provided in the supplementary material.~(See in Sec. \ref{var_backbones}).

\begin{table}
\centering
\scriptsize
\caption{\textbf{Generalization Evaluation} We evaluate the \textit{Masked Logit Nudging} on PIE-Benchmark using two different VAR backbones.}
\label{tab:ablation_backbones}
\begin{tabular}{lccc}
\toprule
\textbf{Backbone}  & \textbf{PSNR↑} & \textbf{LPIPS↓} & \textbf{CLIP↑} \\
\midrule
DDIM (as baseline) & 17.88 & 197.96 & 21.20\\  
Infinity~\cite{han2024infinity} & 27.91 & 44.62 & 21.95 \\
SWITTI~\cite{voronov2024switti}   & {29.70} & {36.50} & {22.72} \\
\bottomrule
\end{tabular}
\end{table}

\section{Conclusion}


We presented Masked Logit Nudging, an architecture-agnostic, inversion-free and prompt-guided approach to image editing for VAR models. Our approach utilises source image token maps to introduce a guidance step that aligns the model’s predictions with these source token maps under the target prompt. Crucially, edits are only applied within spatial masks obtained through a dedicated masking scheme. Furthermore, we introduced a quantization refinement step to correct quantization errors and enhance reconstruction quality. Through extensive evaluation, we demonstrated that our method outperforms VAR-related approaches, achieving comparable or even superior performance to diffusion models while being much faster.
\section{Acknowledgements}
Part of the research leading to these results is funded by the German Research Foundation (DFG) within the project 458972748. The authors would like to thank the foundation for the successful cooperation.

Additionally the authors gratefully acknowledge the scientific support and HPC resources provided by the Erlangen National High Performance Computing Center (NHR@FAU) of the Friedrich-Alexander-Universität Erlangen-Nürnberg (FAU). The hardware is funded by the German Research Foundation (DFG).

{
    \small
    \bibliographystyle{ieeenat_fullname}
    \bibliography{main}
}

\clearpage
\setcounter{page}{1}
\maketitlesupplementary

\section*{Supplementary Material}

This supplementary document provides additional analysis and implementation details for
\emph{Masked Logit Nudging} (MLN). In particular, we include:

\begin{enumerate}
    \item Detailed analysis of the cross-attention–driven edit masks, including quantitative mask--GT comparisons, threshold sensitivity, and layer/head ablations (Sec.~\ref{sec:crossmask}).
    \item Additional comparison and ablations of nudging schedule (Sec.~\ref{decay_schedule}).
    \item Further MLN ablations and hyperparameters~(Sec.~\ref{hyperparameter}).
    \item Extended analysis of quantization errors and the proposed quantization refinement procedure (Secs.~\ref{quant_errors}).
    \item Details and qualitative samples of the reconstruction experiments (Sec.~\ref{sec:inforec}).
    \item {Details and additional qualitative samples of the editing experiments} (Sec.~\ref{sec:qualedits}).
    \item Adapted upscaled PIE-benchmark at 1024px (Sec.~\ref{sr_benchmark}).
    \item Recaptioning for reconstruction experiments at 1024px (Sec.~\ref{recaptioning}).
    \item Additional qualitative editing samples (Sec.~\ref{sec:addqualedits1}).
    \item More ablations (Sec.~\ref{speed_efficiency}).
    \item Failure Analysis (Sec.~\ref{failure_cases}).
\end{enumerate}

\subsection{Cross-attention mask analysis}
\label{sec:supp_mask_analysis}

Our masking mechanism follows the attention-based editing philosophy of DDIM inversion
and P2P~\cite{hertz2023prompt}, but applies it directly to the cross-attention activations
of the VAR transformer, which uses the same multi-head attention structure as GPT-style
models. To extract these activations, we run two short regeneration passes—one with the
source prompt $t_s$ and one with the target prompt $t_t$—from the high-resolution
scales ($s{=}9$ for 512\,px and $s{=}13$ for 1024\,px). The difference between these attention maps yields a spatial relevance map,
which we threshold to obtain the edit mask used by MLN.

\begin{figure*}[t]
    \centering
    \includegraphics[scale=.067]{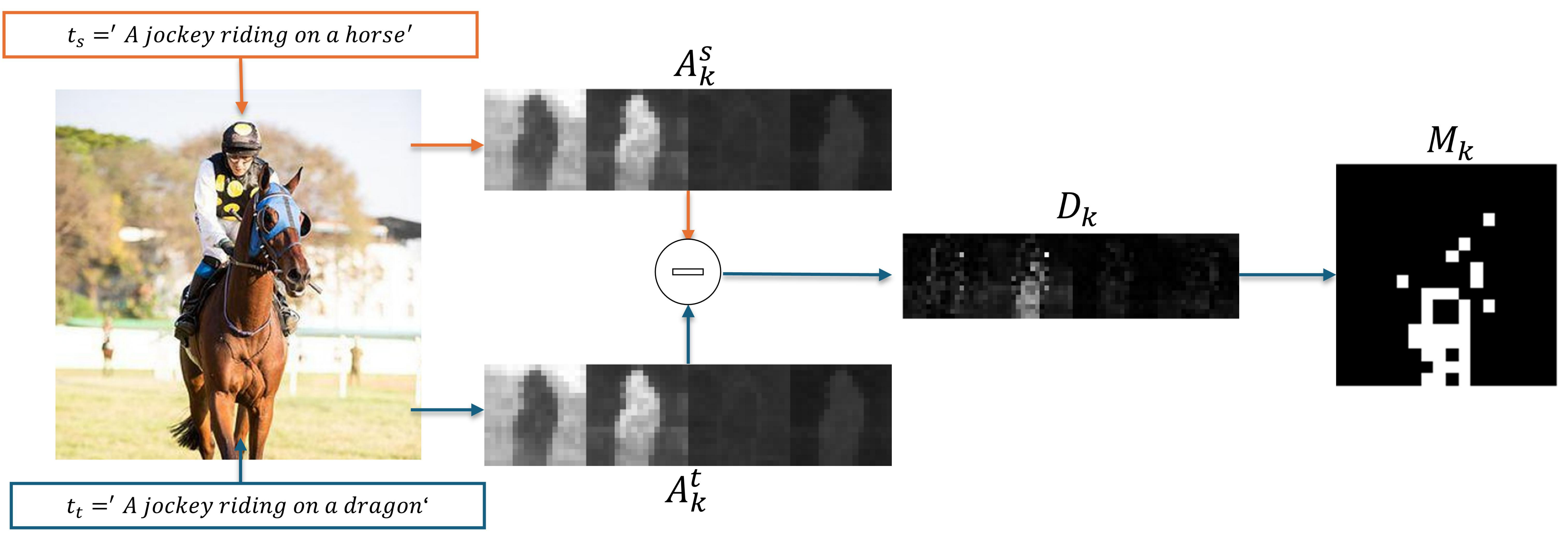}
    \caption{\textbf{Mask construction overview.} Cross-attention differences between
    source and target prompts identify editable regions.}
    \label{fig:masks}
\end{figure*}

In the following we analyze this masking process in detail, focusing on:  
\begin{itemize}
\item how mask-related hyperparameters (regeneration latency, percentile threshold $q$, layer/head selection) affect mask quality and editing performance.
\item how the masks align with the PIE ground-truth edit regions.  

\end{itemize}

Unless otherwise noted, all statistics are computed on the PIE-Benchmark for 512\,px(PIE-512) resolution using the SWITTI backbone~\cite{voronov2024switti}.

\subsubsection{Hyperparameter and latency}
\paragraph{Mask related regeneration latency.}
To extract cross-attention maps, we run regeneration from $s_{\textbf{M}}$ and
record the attention tensors $\mathbf{A}^{s}$ (source prompt) and $\mathbf{A}^{t}$ (target
prompt). The latency below reflects the total time required to compute both
$\mathbf{A}^{s}$ and $\mathbf{A}^{t}$ for a single image. We benchmark this trade-off on
PIE-512 for $s_{\textbf{M}} \in \{5, 6, 7, 8, 9\}$.

\begin{table}[h]
    \centering
    \caption{\textbf{Latency and precision for varying $s$ (512\,px).}}
    \label{tab:mask_s_512_short}
    \scriptsize
    \begin{tabular}{lcc}
        \toprule
        $s_{\textbf{M}}$ & Latency (ms)$\downarrow$ & Precision (\%)$\uparrow$ \\
        \midrule
        5  & 325 & 63 \\
        6  & 122 & 67 \\
        7  & 80  & 69 \\
        8  & 43  & 68 \\
        9* & 20  & 71 \\
        \bottomrule
    \end{tabular}
\end{table}

Latency decreases for larger $s_{\textbf{M}}$ because fewer scale predictions are executed:
when $s_{\textbf{M}}=6$, the model still processes four additional scales, each requiring a full
autoregressive forward pass over increasingly large token grids. Although later scales
contain more tokens, the dominant cost arises from the repeated multi-scale predictions
at earlier stages(since they are sequential and not parallelizable), making shallow $s_{\textbf{M}}$ values substantially slower overall.

Mask precision increases steadily with higher regeneration scale $s_{\textbf{M}}$ and peaks near $s_{\textbf{M}}{=}9$, which
corresponds to almost the full latent resolution ($K{=}10$). Based on this trade-off,
we adopt $s_{\textbf{M}}{=}9$ for 512\,px (and $s_{\textbf{M}}{=}13$ for 1024\,px) in all subsequent experiments.

\paragraph{Threshold sensitivity.}
The binary mask $\mathbf{M}$ is obtained by selecting the top-$q$ percentile of
cross-attention differences, making $q$ the main control over mask sparsity. Low $q$
yields overly small masks, while high $q$ produces masks that spill into the background.

We evaluate $q \in \{60,70,80,90\}$ on PIE-512 and measure mask coverage, IoU with the
ground-truth edit region, and MLN editing quality.

\begin{table}[h]
    \centering
    \caption{\textbf{Effect of threshold $q$ on mask sparsity and editing quality (PIE-512).}}
    \label{tab:mask_threshold}
    \scriptsize
    \begin{tabular}{lccccc}
        \toprule
        $q$ &
        Coverage (\%)$\downarrow$ &
        IoU (\%)$\uparrow$ &
        PSNR$\uparrow$ &
        LPIPS$\downarrow$ &
        CLIP$\uparrow$ \\
        \midrule
        60 & 4.8 & 51 & 29.1 & 0.128 & 0.322 \\
        70 & 7.3 & 57 & 29.4 & 0.121 & 0.331 \\
        80* & 10.6 & 63 & 29.7 & 0.118 & 0.339 \\
        90 & 18.2 & 61 & 29.0 & 0.132 & 0.336 \\
        \bottomrule
    \end{tabular}
\end{table}

Overall, $q{=}80$ offers the best trade-off: it yields the highest IoU and strong editing
performance without unnecessary background changes. We adopt $q=80$
 for 512px and $q=63$ for 1024px.
\paragraph{Layer and head ablations.}

\label{sec:supp_mask_layers}

We aggregate cross-attention maps by averaging over all heads (as also done in
Prompt-to-Prompt~\cite{hertz2023prompt}) and study which transformer decoder blocks
provide the strongest and most stable attention differences. Visually, we observe that
useful attention structure emerges only from layers 3–27: early blocks (0–2) produce
noisy activations, while late blocks (28–29) are overly localized and
inconsistent. The middle layers capture both spatial layout and fine-grained attribute
changes.

To quantify this, we compute masks from different layer ranges on PIE-512 ($q{=}80$) and
measure IoU with ground-truth edit regions together with MLN editing performance.

\begin{table}[h]
    \centering
    \caption{\textbf{Layer-range ablation (PIE-512, $q{=}80$).}  
    Middle blocks yield the most coherent masks and best editing fidelity.}
    \label{tab:mask_layers}
    \scriptsize
    \begin{tabular}{lccc}
        \toprule
        \textbf{Layer Range} &
        \textbf{IoU (\%)}$\uparrow$ &
        \textbf{PSNR}$\uparrow$ &
        \textbf{LPIPS}$\downarrow$ \\
        \midrule
        Early (0–2)       & 41 & 28.6 & 0.151 \\
        Middle (3–27)*     & 63 & 29.7 & 0.118 \\
        Late (28–29)      & 52 & 29.2 & 0.134 \\
        All (0–29)        & 61 & 29.6 & 0.123 \\
        \bottomrule
    \end{tabular}
\end{table}

\begin{figure}[h]
    \centering
    \includegraphics[scale=.2]{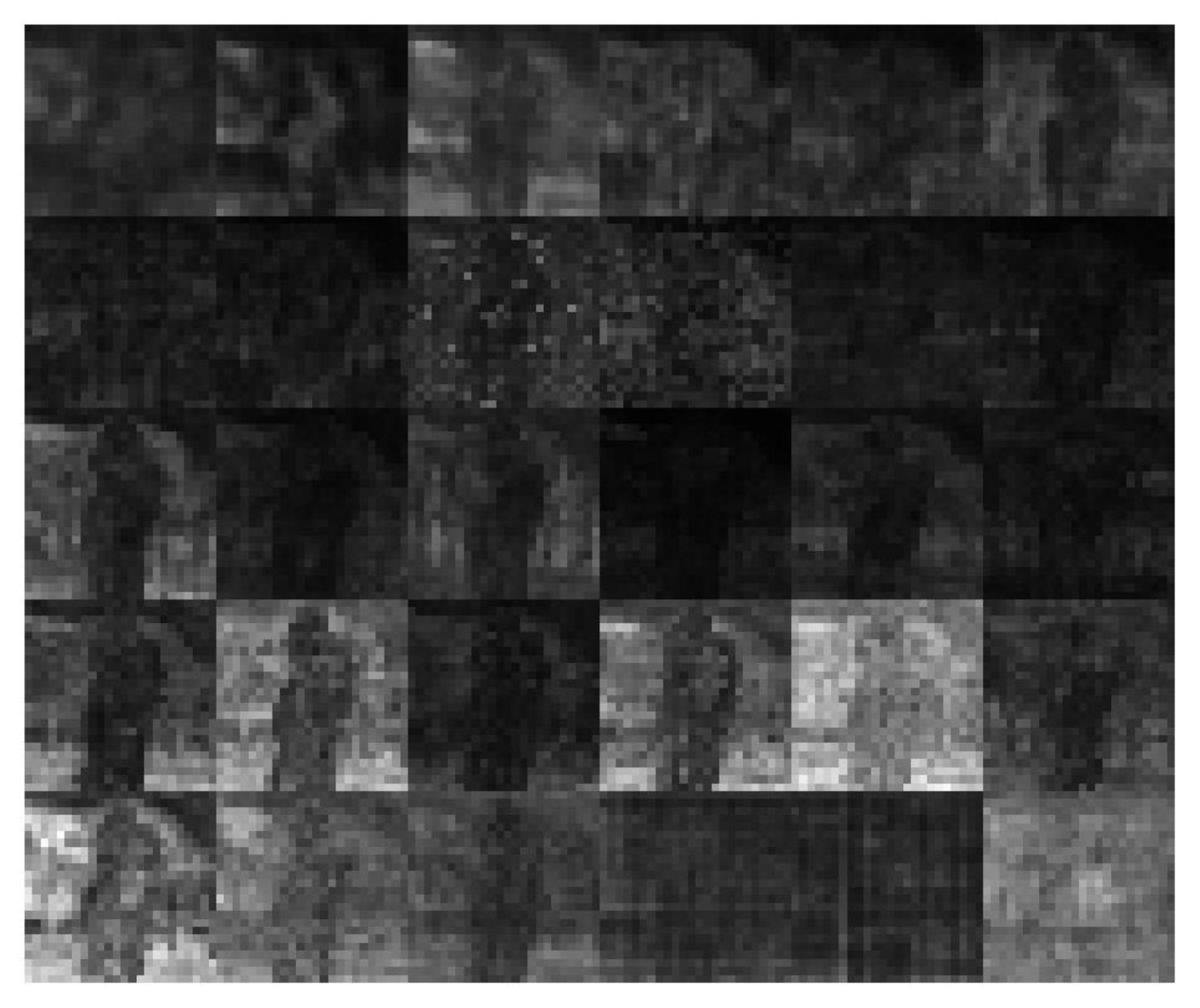}
    \caption{\textbf{Cross-attention maps} for all 30 transformer
    blocks conditioned on the image in fig.~\ref{fig:masks}. Only layers 3–27 yield stable and meaningful masks. Counted left-right from top-bottom.}
    \label{fig:all_block_masks}
\end{figure}

Figure~\ref{fig:all_block_masks} shows the attention-difference maps for
all 30 blocks, illustrating that layers 3–27 provide the cleanest, most semantically
aligned masks. Accordingly, we use blocks 3–27 as the default range in all experiments.

\subsubsection{Mask vs. Ground-Truth Edit Regions}
\label{sec:supp_mask_gt}

We compare our cross-attention–derived masks to the ground-truth edit regions on
PIE-512. While MLN supports explicit masking, it is important to note that
{logit nudging alone already maintains much of the background structure}. Because
the nudging term pulls logits toward the source tokens, the model does not overwrite
large regions as aggressively as plain regeneration~(can also be seen in fig.~\ref{fig:regeneration}). However, without masking(therefore also without
quantization refinement (QR)), the background reconstruction is still worse.

To demonstrate the importance of masking, we compare:
\begin{enumerate}
\item logit nudging without a mask -- no QR.  \item masked regeneration -- no QR.
\item MLN -- with QR.
\end{enumerate}
We measure mask IoU against the PIE ground-truth region and report background fidelity.

\begin{table}[h]
    \centering
    \caption{\textbf{Mask–GT agreement and background fidelity (PIE-512).}   MLN achieves the strongest localization and background preservation.}
    \label{tab:mask_vs_gt}
    \scriptsize
    \setlength{\tabcolsep}{3pt}  

    \begin{tabular}{lcccc}
        \toprule
        \textbf{Method} &
        \textbf{Mask IoU (\%)}$\uparrow$ &
        \textbf{PSNR (bg)}$\uparrow$ &
        \textbf{LPIPS (bg)}$\downarrow$ &
        \textbf{CLIP (edit)}$\uparrow$ \\
        \midrule
        Logit nudging           & -- & 25.8 & 85.2 & 24.4 \\
        Masked regeneration     & 57 & 26.5 &  79.7 & 22.2 \\
        MLN (ours)                             & 63 & 29.7 & 36.5 & 22.72 \\
        \bottomrule
    \end{tabular}
\end{table}
Logit nudging without a mask performs well on edit alignment but fails to
preserve background details, confirming that spatial constraints are essential for
stable reconstructions. Masked regeneration does not improve background  significantly.

Finally, we conclude that applying the mask is beneficial not only for localizing the edit,
but also for {improving reconstruction outside the mask} with the proposed QR.
\label{sec:crossmask}

\subsection{Nudging schedules}
\label{decay_schedule}
\label{sec:supp_nudging_schedules}

Masked Logit Nudging applies a scale-dependent guidance weight $\alpha_k$ at each
VAR scale $k$. For 512\,px images, SWITTI uses $K{=}10$ scales. We found that
the trade-off between edit strength and reconstruction fidelity is best when:

\begin{itemize}
    \item regeneration from $s{=}6$, and
    \item nudging is applied from scale $k \geq 7$, with a decreasing schedule toward the finest scales.
\end{itemize}

\begin{figure}[h]
    \centering
    \begin{tikzpicture}
        \begin{axis}[
            width=0.9\linewidth,
            height=5cm,
            xlabel={Scale $k$},
            ylabel={$\alpha_k$},
            xmin=1, xmax=10,
            ymin=0, ymax=14,
            xtick={1,2,3,4,5,6,7,8,9,10},
            ytick={0,2,4,6,8,10,12,14},
            legend style={at={(0.6,0.27)},anchor=north east,font=\scriptsize},
            grid=both,
            ]
            \addplot[color=red, thick, mark=square*] coordinates {
                (1,12) (2,12) (3,12) (4,12) (5,12) (6,12) (7,12) (8,4) (9,2) (10,0)
            };
            \addlegendentry{Sharp schedule}

            \addplot[color=blue, thick, mark=square*] coordinates {
                (1,12) (2,11.5) (3,11) (4,10) (5,9) (6,8) (7,6) (8,3) (9,1.5) (10,0.5)
            };
            \addlegendentry{Smooth schedule (ours)}

            \draw[dashed, black] (axis cs:7,0) -- (axis cs:7,12);
        \end{axis}
    \end{tikzpicture}
    \caption{\textbf{Nudging schedules at 512\,px.}  
    Both schedules use a cutoff at $k_{\text{cut}}{=}7$ (vertical dashed line). We adopt
    the smooth schedule in all experiments.}
    \label{fig:nudging_schedules}
\end{figure}
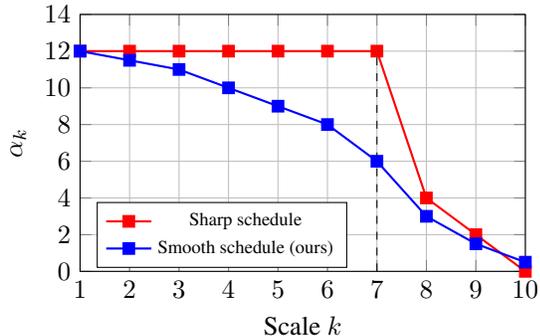

In practice, we use schedules that keep $\alpha_k$ high on early editing
scales (although these scales are not used during MLN, due to regeneration from $s=6$) and then gradually reduce it at high-resolution scales
(to allow fine details without overshooting). Figure~\ref{fig:nudging_schedules} illustrates two representative schedules for 512\,px.

In all reconstruction experiments, we keep the same regeneration scale $s{=}6$. In our experiments we use the smooth schedule.
\paragraph{Nudging cutoff $k$}
\begin{figure*}[t]
\begin{tikzpicture}
\centering
  \node[inner sep=0] (image) {
    \includegraphics[scale=.067]{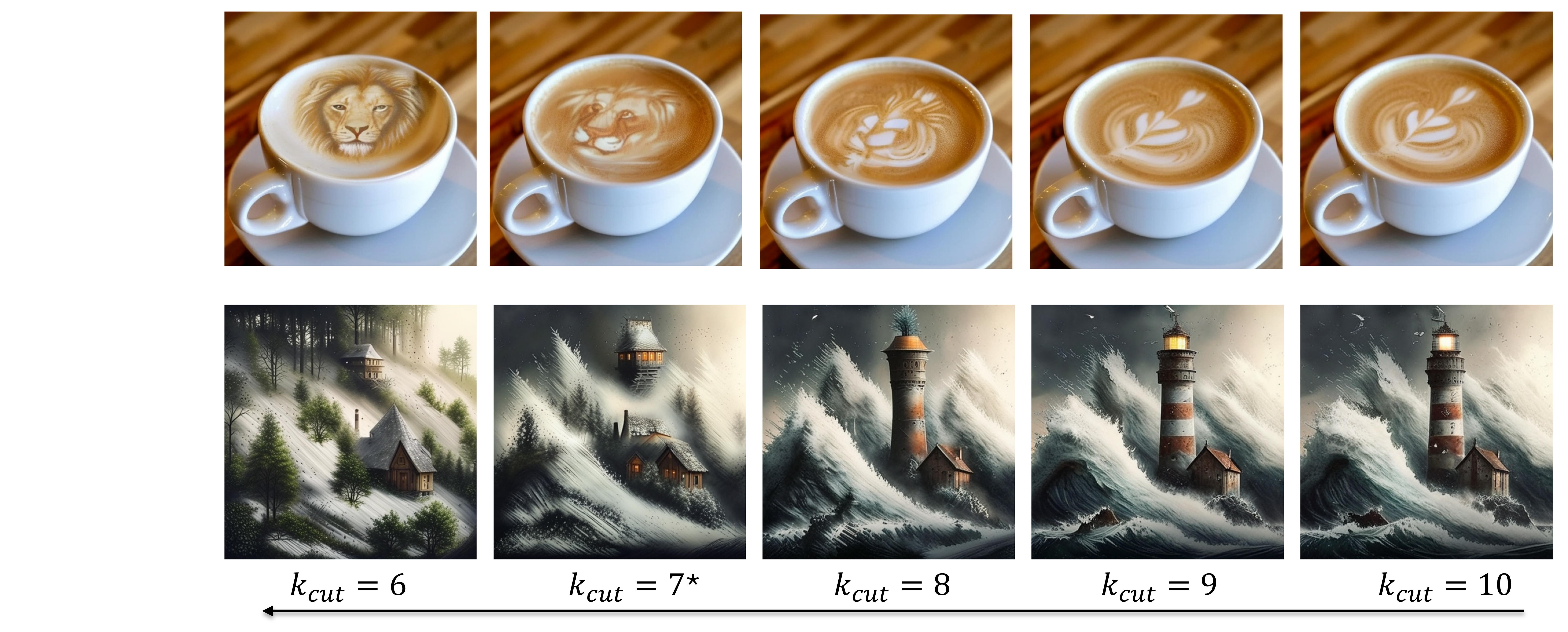}
  };

  \begin{scope}[shift={(image.south west)}, x=1cm, y=1cm]

    \usetikzlibrary{positioning}

    \tikzset{
      colhead/.style={
        font=\fontsize{7pt}{5pt}\editfont,
      },
      rowlbl/.style={
        font=\normalsize,
      }
    }

    \node[colhead,  anchor=north] at (1,5)
      {"[\sout{tulip}$\rightarrow$\contour{black}{lion}]".};
    \node[colhead,  anchor=north, align=center] at (1,2.5)
      {"[\sout{sea} $\rightarrow$\contour{black}{forest}] \\and house".};
  \end{scope}
\end{tikzpicture}
\caption{\textbf{Nudging cutoff $k_{cut}$.} Higher $k_{cut}$ preserves more content to the original image (seen at $k_{cut}=10$). Importantly the upper example utilizes a mask~(MLN) to keep edits from the background, while the lower example only uses logit-nudging.}

\label{fig:nudgingcut}
\end{figure*}

Additionally we ablate different cutoff scales $k_{\text{cut}}$ on PIE-512 using the smooth schedule (see Tab.~\ref{tab:nudging_cutoff}). Evaluations include background PSNR,
background LPIPS, and CLIP alignment in the edited region. 
Visual samples are shown in fig.~\ref{fig:nudgingcut}.

\begin{table}[h]
    \centering
    \caption{\textbf{Ablation over cutoff scale $k_{\text{cut}}$ (PIE-512, smooth schedule).}}
    \label{tab:nudging_cutoff}
    \scriptsize
    \begin{tabular}{lcc}
        \toprule
        $k_{\text{cut}}$ &
        PSNR (bg)$\uparrow$ &
        CLIP (edit)$\uparrow$ \\
        \midrule
        5  & 21.4 & 26.1 \\
        6  & 22.0 & 24.4 \\
        7* & 24.7 & 22.7 \\
        8  & 25.9 & 20.5 \\
        9  & 27.9 & 19.3 \\
        \bottomrule
    \end{tabular}
\end{table}

$k_{\text{cut}}{=}7$ provides the best trade-off between background
fidelity (highest PSNR, lowest LPIPS) and edit strength (highest CLIP). Later cutoffs
over-constrain fine scales and weaken edits, while earlier cutoffs allow excessive nudging
at high resolution and degrade background preservation.
\subsection{MLN ablations}
\label{hyperparameter}

\subsubsection{Component-wise Ablations}
\label{sec:supp_component_ablation}

We evaluate the contribution of each MLN component on PIE-512.
The three components analyzed are:  
\begin{itemize}
    \item \textbf{Logit Nudging (LN)} for semantic steering,  
    \item \textbf{Cross-Attention Masking (Mask)} for spatial localization, and 
    \item \textbf{Quantization Refinement (QR)} for restoring background regions.
\end{itemize}

\begin{table}[h] \centering \caption{\textbf{Component-wise ablations on PIE-512.} Columns indicate which components are enabled in each variant.} \label{tab:component_ablation} \scriptsize \begin{tabular}{ccccccc} \toprule \textbf{LN} & \textbf{Mask} & \textbf{QR} & \textbf{PSNR (bg)}$\uparrow$ & \textbf{LPIPS (bg)}$\downarrow$ & \textbf{CLIP (edit)}$\uparrow$ \\ \midrule \checkmark & & & 23.4 & 134.2 & 25.20 \\ & \checkmark & & 24.5 & 120.7 & 21.20 \\ \checkmark & \checkmark & & 27.6 & 102.3 & 22.32 \\  \checkmark & \checkmark & \checkmark & 29.70 & 36.50 & 22.72 \\ \bottomrule \end{tabular} \end{table}

LN enhances edit strength by providing semantic steering, but without RQ background regions are not preserved as good.
Masking alone offers almost no improvements by spatially restricting edits, as shown in sec.~\ref{sec:supp_mask_gt} it performs only slightly better than LN.
The combination of LN and masking yields the largest performance gains, enabling edits that are both semantically aligned and spatially well-localized, however background preservation still  suffers.
Incorporating all three components produces the most robust results overall.

\paragraph{Bakground preservation weight $\beta$}

\begin{figure}[h]
\centering
\begin{tikzpicture}

    \begin{axis}[
        width=0.9\linewidth,
        height=6cm,
        xlabel={$\beta$},
        xtick={0,2,4,6,8,10,12,14,16},
        xmin=0,xmax=16,
        ymin=25, ymax=30,
        ylabel={PSNR (bg) [dB]},
        ytick={26,27,28,29,30,31},
        legend style={at={(0.21,0.01)},anchor=south},
        axis y line*=left,
        axis x line*=bottom,
        grid=both
    ]

    \addplot[color=blue, thick, mark=*] coordinates {
        (0,26.1)
        (2,26.9)
        (4,27.8)
        (6,28.6)
        (8,29.3)
        (10,29.5)
        (12,29.7)
        (14,29.9)
        (16,30)
    };
    \addlegendentry{PSNR (bg)$\uparrow$}

    \end{axis}

    \begin{axis}[
        width=0.9\linewidth,
        height=6cm,
        xtick={0,2,4,6,8,10,12,14,16},
        xmin=0, xmax=16,
        ylabel={CLIP (edit)},
        ytick={20.5,21,21.5,22,22.5,23},
        ymin=20.5, ymax=23,
        axis y line*=right,
        axis x line=none,
        legend style={at={(0.702,0.01)},anchor=south},
        grid=both
    ]

    \addplot[color=red, thick, dashed, mark=square*] coordinates {
        (0,22.5)
        (2,22.56)
        (4,22.4)
        (6,22.6)
        (8,22.7)
        (10,22.7)
        (12,22.9)
        (14,22.1)
        (16,21.0)
    };
    \addlegendentry{CLIP (edit)$\uparrow$}

    \end{axis}

\end{tikzpicture}
\caption{\textbf{Ablation over $\beta$.}  
PSNR improves up to $\beta{=}12$. CLIP remains stable until $\beta{=}14$ and then saturates.}
\label{fig:beta_ablation}
\end{figure}
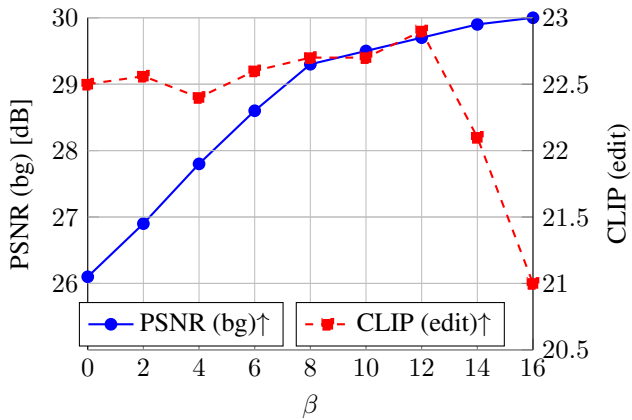

The weight $\beta$ controls the strength of background preservation during MLN~(this can be seen in fig.~\ref{fig:beta_ablation}).  
A larger $\beta$ penalizes deviations outside the mask, improving reconstruction but
potentially weakening the edit if set too high.  
We vary $\beta \in \{0,2,4,\dots,16\}$ on PIE-512 and measure
PSNR (background region) and CLIP similarity (edit region).  
PSNR increases steadily up to $\beta{=}12$ and then saturates. CLIP improves until $\beta{=}14$, after which it saturates.  
We therefore use $\beta{=}12$ as the default value.

\paragraph{Regenereation scale $s$ during MLN}
\label{sec:supp_s_ablation}

MLN begins editing from an intermediate VAR scale $s$, reusing source tokens for all
lower scales and applying nudging only at higher scales~(see Tab.~\ref{tab:mln_s_ablation}). Choosing $s$ therefore
determines the trade-off between preserving global structure and allowing sufficient room for edits to form.  
We evaluate $s\in\{4,5,6,7,8\}$ on PIE-512 using identical settings ($q{=}80$, $\beta{=}12$).

\begin{table}[h]
    \centering
    \caption{\textbf{Ablation of MLN starting scale $s$ (PIE-512)}
    Increasing $s$ improves background fidelity but weakens edits; $s{=}6$ yields the
    best compromise. Latency measures the time required to run the regeneration/MLN
    forward pass starting from scale $s$.}
    \scriptsize
    \label{tab:mln_s_ablation}
    \begin{tabular}{lcccc}
        \toprule
        $s$ &
        PSNR (bg)$\uparrow$ &
        LPIPS (bg)$\downarrow$ &
        CLIP (edit)$\uparrow$ &
        Latency (ms)$\downarrow$ \\
        \midrule
        4 & 22.1 & 180.0 & 23.2& 1210 \\
        5 & 26.4 & 50.3 & 22.6 & 962 \\
        6* & {29.3} & {36.5} & {22.7} & 800 \\
        7 & 30.8 & 25.2 & 20.8 & 413 \\
        8 & 31.0 & 112.2 & 18.1 & 84 \\
        \bottomrule
    \end{tabular}
\end{table}

Background fidelity improves monotonically with increasing $s$, while CLIP alignment
begins to drop once too few scales remain for meaningful edits.  
The best overall balance is obtained at \textbf{$s{=}6$}, which we adopt as the default
for all 512\,px experiments (and $s{=}10$ for 1024\,px).

\begin{figure*}[h]
    \centering
    \includegraphics[scale=.076]{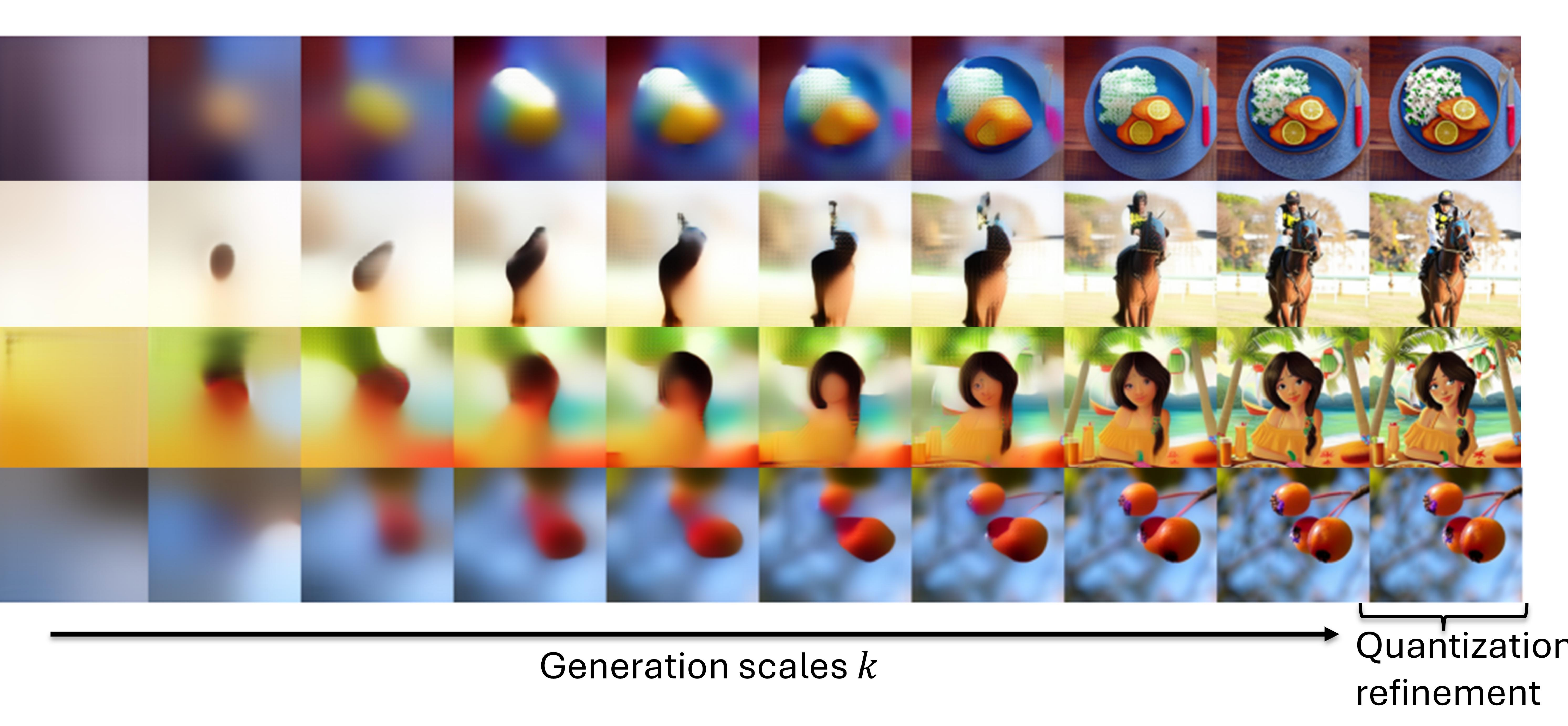}
    \caption{\textbf{Generation scales.} Visual comparison of SWITTI generation at all scales $k$. As $k$ increases, images get more high-frequency details. We choose $k=s=7$ as regeneration scale.}
    \label{fig:s_scale_visual}
\end{figure*}
\paragraph{Sampling Hyperparameters (CFG Schedule)}
\label{sec:supp_sampling}
Since early VAR scales are structurally important and later scales contain high-frequency appearance details, we activate CFG-style guidance only in a narrow mid-scale band.

Concretely, we use the following schedule:

\begin{itemize}
    \item CFG sampling is \textbf{enabled starting at scale $k=2$}, once the global layout has been established.
    \item CFG sampling is \textbf{disabled again at scale $K-2$} (i.e., $k=8$ for $K{=}10$ at 512\,px).
    \item Outside this range, we perform standard sampling without CFG adjustment.
    \item The same schedule is adopted for 1024\,px with $K{=}14$, i.e., CFG active from $k{=}2$ to $k{=}12$.
\end{itemize}

This mid-band CFG improves prompt alignment without destabilizing fine-scale token predictions,
and we observe no benefit from applying CFG at the very first or very last scales.

\subsection{Extended analysis of quantization errors and quantization refinement}
\label{quant_errors}
\label{quant_errors}

\begin{figure}[h]
    \centering
    \includegraphics[scale=.05]{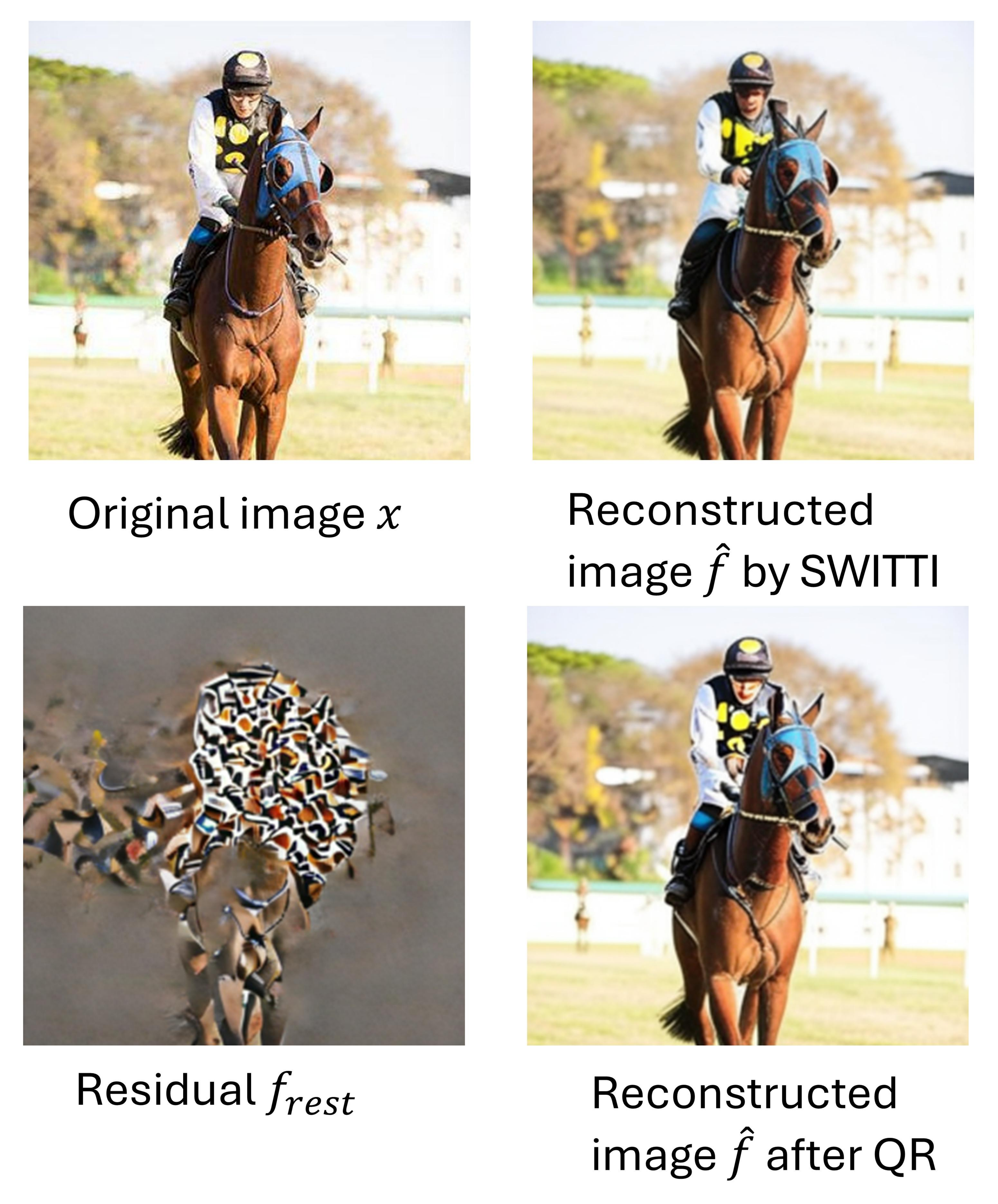}
    \caption{\textbf{Visualization of quantization refinement}. Residuals accumulate outside the codebook manifold, causing the default SWITTI reconstruction to make mistakes.}
    \label{fig:quant_err_vis}
\end{figure}

\paragraph{Quantization Errors in VAR.}
During encoding, each continuous feature map $\mathbf{f}$ is approximated by discrete
codebook vectors $\{\mathbf{f}_k\}_{k=1}^K$. This introduces a residual error
\[
\mathbf{f}_{\text{rest}}
    \;=\;
    \mathbf{f}
    \;-\;
    \sum_{k=1}^K \mathbf{f}_k,
\]
which accumulates across scales and causes visible distortions after decoding—
typically slight color drifts, softened textures, and structural inconsistencies.
We visualize these errors in Fig.~\ref{fig:quant_err_vis}, where
the reconstructed image without refinement  (fig.~\ref{fig:quant_err_vis},top right) deviates from the VQ-manifold, resulting in images that are reconstructed badly.

\paragraph{Quantization Refinement - Mathematical Perspective.}

Because $\mathbf{f}_{\mathrm{rest}}$ generally lies {off} the codebook
manifold spanned by the embeddings $\mathbf{C}=\{c_1,\dots,c_V\}$, adding it directly to $\hat{\mathbf{f}}$ produces severe artifacts(fig.~\ref{fig:quant_err_vis}, bottom left).  
We therefore {project} the residual back into the codebook space before applying
it as a correction.

For each iteration $j$, we treat the residual 
$\mathbf{f}_{\mathrm{rest}}^{(j)}\in\mathbb{R}^C$ as
continuous observations and compute soft assignment weights
\[
w_i^{(j)}
 \;=\;
\frac{
    \exp\!\left(\langle \mathbf{f}_{\mathrm{rest}}^{(j)},\, c_i\rangle / \tau\right)
}{
    \sum_{k=1}^V \exp\!\left(\langle \mathbf{f}_{\mathrm{rest}}^{(j)},\, c_k\rangle / \tau\right)
},
\]
where $\tau$ is a temperature controlling assignment sharpness.  
The projected residual is then
\[
\mathbf{f}_{\mathrm{proj}}^{(j)}
\;=\;
\sum_{i=1}^V w_i^{(j)}\, c_i,
\]
which lies exactly in the space of the codebook embeddings $\mathbf{c_i}$.

We then update the reconstruction using a step size $\alpha$,
\[
\hat{\mathbf{f}}^{(j+1)}
\;=\;
\hat{\mathbf{f}}^{(j)} \;+\; \alpha\,\mathbf{f}_{\mathrm{proj}}^{(j)},
\]
and only apply the correction outside the edit mask $\mathbf{m}$:
\[
\hat{\mathbf{f}}^{(j+1)}
\;=\;
\hat{\mathbf{f}}^{(j)} + \alpha\,(\bar{\mathbf{M}})\odot \mathbf{f}_{\mathrm{proj}}^{(j)}.
\]

\paragraph{Iterative refinement}
The off-manifold residual $\mathbf{f}_{\mathrm{rest}}$ contains components that cannot be removed in a single projection step:  
each projection eliminates only the portion explainable by the codebook, while the remaining off-manifold residual changes shape after every update.  Thus, repeating the projection–update cycle gradually decreases the residual norm,
\[
\|\mathbf{f}_{\mathrm{rest}}^{(j)}\|_2 \downarrow,
\]
until the correction becomes negligible, yielding a reconstruction closer to the original $\mathbf{f}$ while keeping the edited region intact. In practice we use $j=5$ iteration and $\tau=0.2$ at 512px and $j=3$ with $\tau=0.8$ at 1024px. The pseudocode can be seen in algorithm \ref{alg:soft_projection}.

\begin{algorithm}[t]
\caption{Quantization refinement}
\label{alg:soft_projection}
\begin{algorithmic}[1]
\Require Encoded features $\mathbf{f}$,initial reconstruction $\hat{\mathbf{f}}^{(0)}$, codebook embeddings $\mathbf{C}$, iterations $T$, temperature $\tau$, step size $\alpha$, tolerance $\varepsilon$,mask $\mathbf{M}$.

\State $\hat{\mathbf{f}} \leftarrow \hat{\mathbf{f}}^{(0)}$
\State $\mathbf{f}_{\text{out}} \leftarrow \hat{\mathbf{f}}^{(0)}$ \myComment{Final refined features}
\For{$j = 1, 2, \dots, T$}
    \State $\mathbf{f}_{\text{rest}} \leftarrow \mathbf{f} - \hat{\mathbf{f}}$ \myComment{Residual}
    \State $r \leftarrow \|\mathbf{f}_{\text{rest}}\|_2$ (e.g.\ mean $\ell_2$ norm)
    \If{$r < \varepsilon$}
        \State \textbf{break} \myComment{Early stopping}
    \EndIf
    \State Reshape $\mathbf{f}_{\text{rest}}$ to $\mathbf{Z} \in \mathbb{R}^{N \times C}$, where $N = BHW$
    \State $\mathbf{S} \leftarrow \mathbf{Z} \mathbf{C}^\top \in \mathbb{R}^{N \times V}$ \myComment{Similarities to codebook}
    \State $\mathbf{W} \leftarrow \mathrm{softmax}(\mathbf{S} / \tau, \text{dim}=V)$ \myComment{Soft assignments}
    \State $\mathbf{Z}_{\text{proj}} \leftarrow \mathbf{W} \mathbf{C} \in \mathbb{R}^{N \times C}$ \myComment{Projection to codebook space}
    \State Reshape $\mathbf{Z}_{\text{proj}}$ back to $\mathbf{f}_{\text{rest}}^{\text{proj}} \in \mathbb{R}^{B \times C \times H \times W}$
    \State $\hat{\mathbf{f}} \leftarrow \hat{\mathbf{f}} + \alpha\, \mathbf{f}_{\text{rest}}^{\text{proj}}$
    \State $\mathbf{f}_{\text{out}} \leftarrow \mathbf{f}_{\text{out}} + \alpha\, \mathbf{f}_{\text{rest}}^{\text{proj}} \odot (\bar{\mathbf{M}})$ \myComment{Refine outside edit mask}
\EndFor
\State \textbf{Output:} $\mathbf{f}_{\text{out}}$
\end{algorithmic}
\end{algorithm}
\paragraph{Latency.}
The quantization–refinement step is computationally negligible compared to the VAR forward pass. Each iteration involves only matrix multiplications with the codebook
($V{\times}C$) and per-pixel softmax operations, both of which are highly optimized on
modern GPUs. In practice, running $10$ refinement iterations adds {less than $\mathbf{10}$\,ms} of overhead for both $512$ and $1024$ resolutions, making the
procedure effectively free relative to the overall editing pipeline. As a result, the
refinement can be applied by default without compromising real-time editing speed.

\label{quant_refinement}
\subsection{Details and qualitative samples of reconstuction experiments}
\label{sec:inforec}
\label{sec:supp_reconstruction_details}
\subsubsection{Reconstruction methods}
To evaluate reconstruction fidelity in the zero-edit setting (i.e., source and target prompts identical), 
we benchmark MLN against a set of diffusion-based baselines.  
This subsection summarizes all methods included in the reconstruction experiments, together with the 
backbones and sampling configurations used in our evaluation.  
All experiments are performed on COCO (512\,px) and the curated 
OpenImages subset (1024\,px) using the official evaluation splits. In the experiments we always relied on the source code provided by the authors, if not otherwise indicated, whenever the source code was available we tested the provided configurations and chose the best one with respect to reconstruction performance.

\vspace{0.5em}
\paragraph{Overview of evaluated methods at 512px.}
We provide an overview with details about the methods in Tab.~\ref{tab:reconstruction_methods}.
\begin{table*}[t]
\centering
\caption{\textbf{Methods used in the 512px reconstruction experiments.} 
For each method we list the backbone model, the reconstruction procedure, 
and the exact settings used in our evaluation.}
\label{tab:reconstruction_methods}
\begin{tabular}{p{2.5cm} p{2.3cm} p{2.8cm} p{3.6cm}}
\toprule
\textbf{Method} & \textbf{Backbone} & \textbf{Procedure} & \textbf{Settings} \\
\midrule

Default SWITTI (baseline) 
& SWITTI~\cite{voronov2024switti} 
& VAR (no refinement) 
& $s{=}6$ (512px), CFG enabled for scales $2$–$8$, default SWITTI sampling \\

TurboEdit~\cite{deutch2024turboedit}
& SDXL~\cite{podellsdxl}
& Diffusion 
& 4 denoising steps \\

ReNoise~\cite{garibi2024renoise}
& SD2.1~\cite{rombach2022high}
& Diffusion
& 50 inversion + 50 inference steps \\

PnP~\cite{ju2023direct}
& SD1.4~\cite{rombach2022high}
& Diffusion
& 50 inference steps \\

Ledits++~\cite{brack2024ledits++}
& SD1.5~\cite{rombach2022high}
& Diffusion 
& 50 inversion steps \\

RF-Inversion~\cite{routsemantic}
& FLUX-1 dev~\cite{flux2024}
& Rectified Flow 
& 28 inversion steps \\

EditFriendly~\cite{huberman2024edit}
& SD1.5~\cite{rombach2022high}
& Diffusion 
& 100 inversion steps \\
\bottomrule
\end{tabular}
\end{table*}

\vspace{0.5em}

\paragraph{Overview of evaluated methods at 1024px.}
We provide an overview with details about the methods in Tab.~\ref{tab:reconstruction_methods2}.

\begin{table*}[t]
\centering
\caption{\textbf{Methods used in the 1024px reconstruction experiments.} 
For each method we list the backbone model, the reconstruction procedure, 
and the exact settings used in our evaluation.}
\label{tab:reconstruction_methods2}
\begin{tabular}{p{2.5cm} p{2.3cm} p{2.8cm} p{3.6cm}}
\toprule
\textbf{Method} & \textbf{Backbone} & \textbf{Procedure} & \textbf{Settings} \\
\midrule

Default SWITTI (baseline) 
& SWITTI~\cite{voronov2024switti} 
& VAR (no refinement) 
& $s{=}8$ (1024px), CFG enabled for scales $2$–$12$, default SWITTI sampling \\

TurboEdit~\cite{deutch2024turboedit}
& SDXL~\cite{podellsdxl}
& Diffusion 
& 4 denoising steps \\

ReNoise~\cite{garibi2024renoise}
& SDXL~\cite{podellsdxl}
& Diffusion 
& 50 inversion + 50 inference steps \\

PnP~\cite{ju2023direct}
& SD1.4~\cite{rombach2022high}
& Diffusion 
& 50 inference steps \\

Ledits++~\cite{brack2024ledits++}
& SDXL~\cite{podellsdxl}
& Diffusion 
& 50 inversion steps \\

RF-Inversion~\cite{routsemantic}
& FLUX-1 dev~\cite{flux2024}
& Rectified Flow 
& 28 inversion steps \\

\bottomrule
\end{tabular}
\end{table*}
\subsubsection{Additional Reconstruction Results}
\label{sec:supp_extra_recon}

In this section, we provide additional qualitative and quantitative results for the
reconstruction experiments at 512\,px and 1024\,px. Unless otherwise stated, all results
use the SWITTI backbone with our default settings.

\begin{figure*}[h]
    \centering
    \includegraphics[scale=.09]{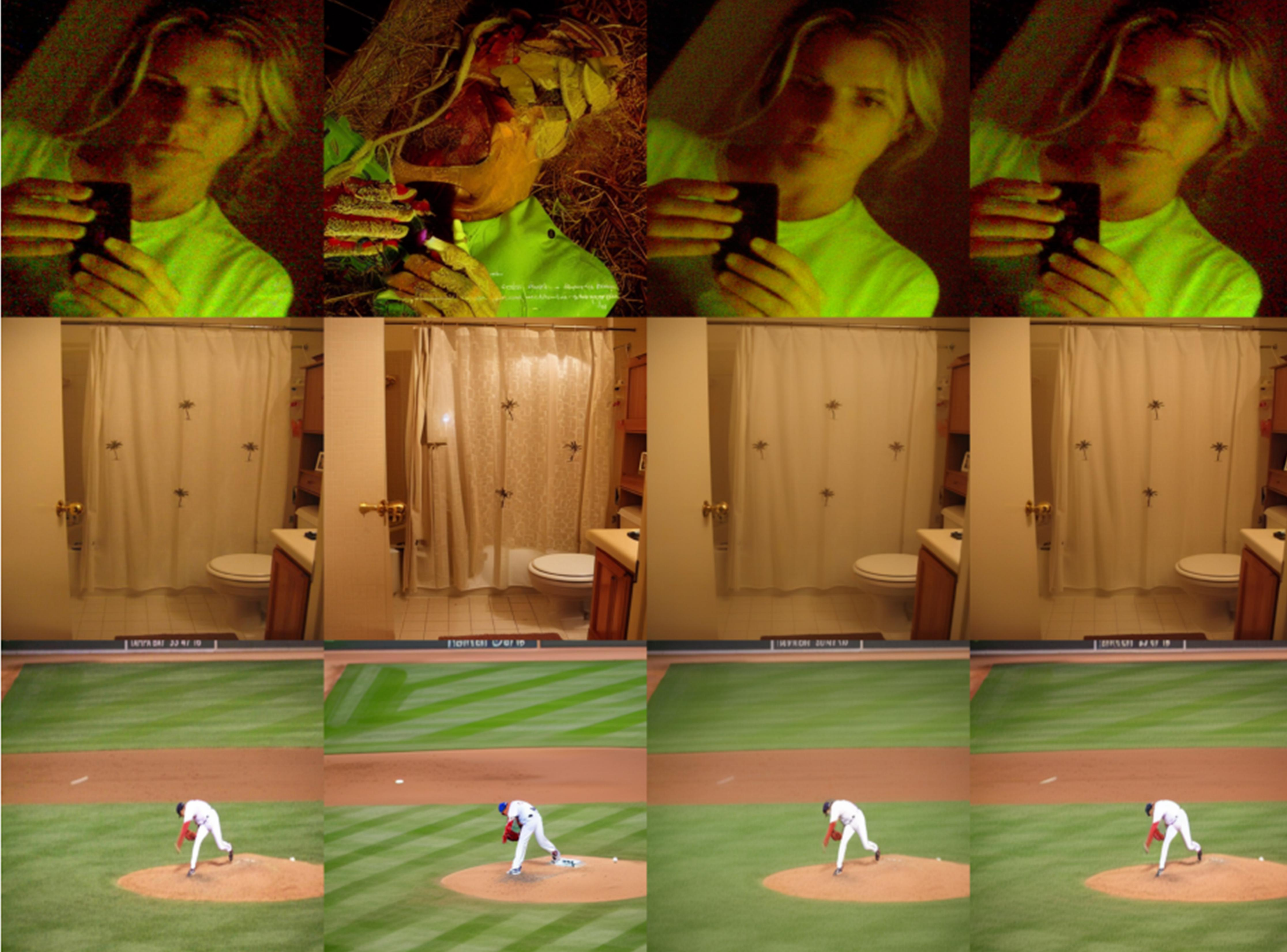}
    \caption{\textbf{Reconstructions on COCO-512.} From left to right:
    input image, SWITTI w/o QR, TurboEdit, SWITTI w QR. QR reduces quantization artifacts
    and preserves local details compared to the baseline and diffusion-based reconstruction.}
    \label{fig:recon_512_qual}
\end{figure*}

\paragraph{512\,px Reconstructions (COCO).}
We first compare reconstructions at 512\,px resolution for three variants:
\begin{itemize}
    \item SWITTI with quantization refinement (QR),  
    \item SWITTI without QR, and  
    \item TurboEdit~\cite{deutch2024turboedit}.  
\end{itemize}
Reconstructions are shown in fig.~\ref{fig:recon_512_qual}.

\begin{figure}[h]
    \centering
    \includegraphics[scale=.05]{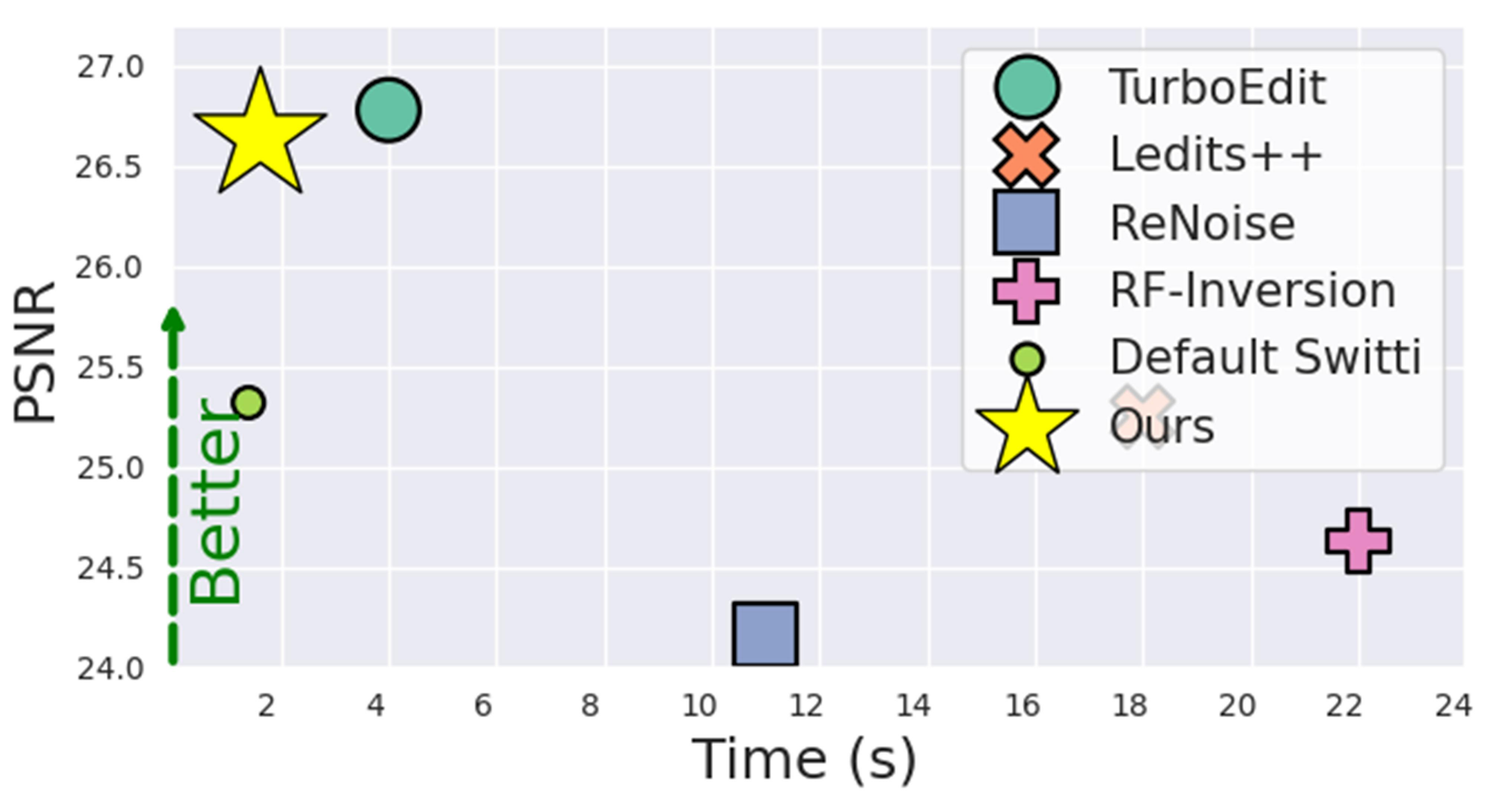}
    \caption{\textbf{PSNR of different reconstruction methods at 1024\,px.} SWITTI with QR achieves the highest PSNR, outperforming both the
    non-refined SWITTI baseline and diffusion/flow-based methods.}
    \label{fig:recon_1024_psnr}
\end{figure}

\paragraph{1024\,px PSNR Comparison (OpenImages).}
At 1024\,px, we report a method-level comparison in terms of PSNR over the OpenImages
subset, including SWITTI w/ and w/o QR and all diffusion/flow baselines used in the main
paper (TurboEdit, ReNoise, PnP, Ledits++, RF-Inversion, EditFriendly). The quantitative results can be seen in fig.~\ref{fig:recon_1024_psnr}.

\begin{figure*}[h]
    \centering
    \includegraphics[scale=.6]{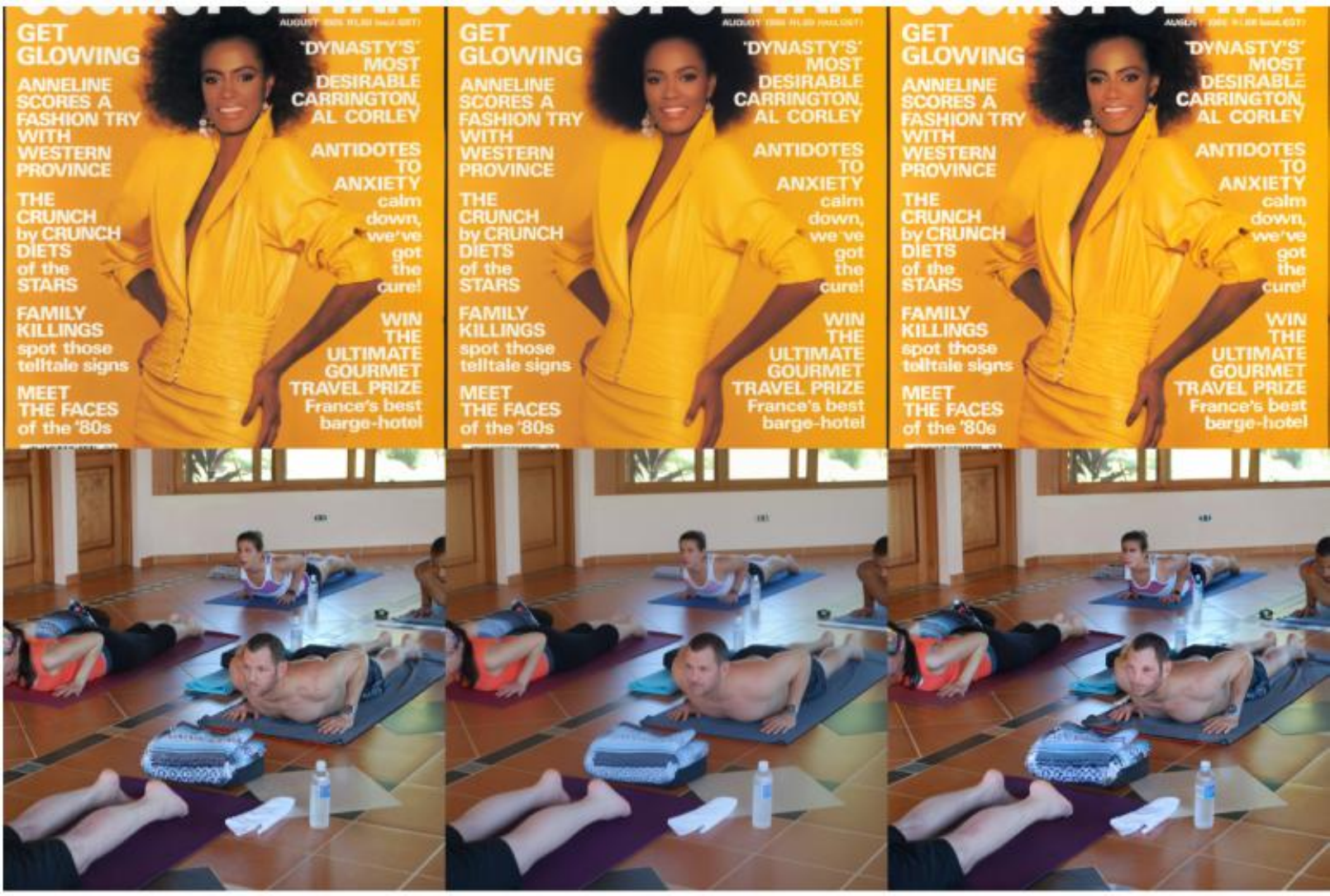}

    \caption{\textbf{ Qualitative reconstruction    comparison at 1024\,px.} From left to right:
    input image, TurboEdit, SWITTI w/o QR, SWITTI w/ QR. Quantization refinement yields
    visibly sharper reconstructions and fewer artifacts, especially in textures and edges.}
    \label{fig:recon_1024_qual}
\end{figure*}

\paragraph{1024\,px Qualitative Comparison of QR.}
Finally, we provide a qualitative comparison at 1024\,px between:
\begin{enumerate}
\item  TurboEdit,  
\item SWITTI without QR, and  
\item  SWITTI with QR.  
\end{enumerate}
This visualization highlights how QR specifically reduces blocky artifacts and restores
sharpness in high-frequency regions without introducing over-smoothing~(see fig.~\ref{fig:recon_1024_qual}).

\subsection{Details and qualitative samples of editing experiments}
\label{sec:qualedits}

In our PIE-Bench editing experiments, we evaluate our method against recent diffusion-based and flow-based baselines. All baseline results reported in this paper were reproduced using the official code released by the respective authors, executed with the recommended default hyperparameters documented in their repositories.

Whenever multiple configuration presets or parameter options were provided, we evaluated the available variants and report the best-performing setting for each method for fair comparison. Tab.~\ref{tab:editing_settings} contrains all the method configurations for the comparison of Tab.~\ref{tab:piebench} in the main paper\footnote{For discrete autoregressive approaches, official codebases were not publicly available. However, we reproduced the reported results for VARIN~\cite{dao2025discrete} and AREdit~\cite{wang2025training} following the descriptions in their papers. Performance metrics closely match the reported values, while efficiency numbers are recomputed based on our hardware setup (NVIDIA A6000).
}.

Tab.~\ref{tab:editing_settings_1024} lists all method configurations from Tab.~\ref{tab:pie_1024_small}.

\begin{table*}[t]
\centering
\caption{Overview of evaluated editing methods, their backbone models, and the specific method configurations used in our experiments.}
\vspace{0.4em}
\label{tab:editing_settings}
\begin{tabular}{l l p{6.2cm}}
\toprule
\textbf{Method} & \textbf{Backbone} & \textbf{Method Settings} \\
\midrule
DDIM~\cite{songdenoising} & SD1.4~\cite{rombach2022high} & DDIM inversion with Prompt-to-Prompt cross-attention control. \\

ReNoise~\cite{garibi2024renoise} & SDXL~\cite{podellsdxl} & 50 inversion steps + 50 inference steps. \\

LEdits++~\cite{brack2024ledits++} & SD1.5~\cite{podellsdxl} & 50-step diffusion inversion. \\

EditFriendly~\cite{huberman2024edit} & SD1.5~\cite{rombach2022high} & DDPM inversion with 100 denoising steps. \\

PnP~\cite{ju2023direct} & SD1.4~\cite{rombach2022high} & 50-step diffusion inversion. \\

TurboEdit~\cite{deutch2024turboedit} & SDXL-Turbo~\cite{podellsdxl} & 4 denoising steps using SDXL-Turbo. \\
\bottomrule
\end{tabular}
\end{table*}

\begin{table*}[t]
\centering
\caption{Overview of evaluated editing methods, their backbone models, and the specific method configurations used in our 1024px PIE experiments.}
\vspace{0.4em}
\label{tab:editing_settings_1024}
\begin{tabular}{l l p{6.2cm}}
\toprule
\textbf{Method} & \textbf{Backbone} & \textbf{Method Settings} \\
\midrule
PnP~\cite{ju2023direct} & SD1.4~\cite{rombach2022high} & 50-step diffusion inversion. \\

LEdits++~\cite{brack2024ledits++} & SDXL~\cite{podellsdxl} & 50-step diffusion inversion. \\

ReNoise~\cite{garibi2024renoise} & SDXL~\cite{podellsdxl} & 50 inversion + 50 inference steps. \\

TurboEdit~\cite{deutch2024turboedit} & SDXL~\cite{podellsdxl} & 4 denoising/editing steps. \\

RF-Inversion~\cite{routsemantic} & Flux~1.dev~\cite{flux2024} & 28-step rectified-flow inversion. \\
\bottomrule
\end{tabular}
\end{table*}
\paragraph{Mask deactivation during style edits}
As discussed in Sec.~\ref{sec:implementation_detail}, for style–transfer scenarios in the PIE benchmark we disable the masking mechanism entirely, both at 512px and 1024px resolution. Concretely, for all samples belonging to the category '9\_change\_style', we enforce full editing on the entire image by manually setting the editing mask to one, i.e., $\mathbf{M}_k = \mathbf{1}$ for all scales $k$. This ensures that stylistic transformations are applied globally, which is necessary because style edits typically require modifications across the entire image rather than localized changes.
\paragraph{Additional samples at 1024px.}
\begin{figure*}[t]
\centering
\begin{tikzpicture}
  \node[inner sep=0] (image) {
    \includegraphics[scale=.09]{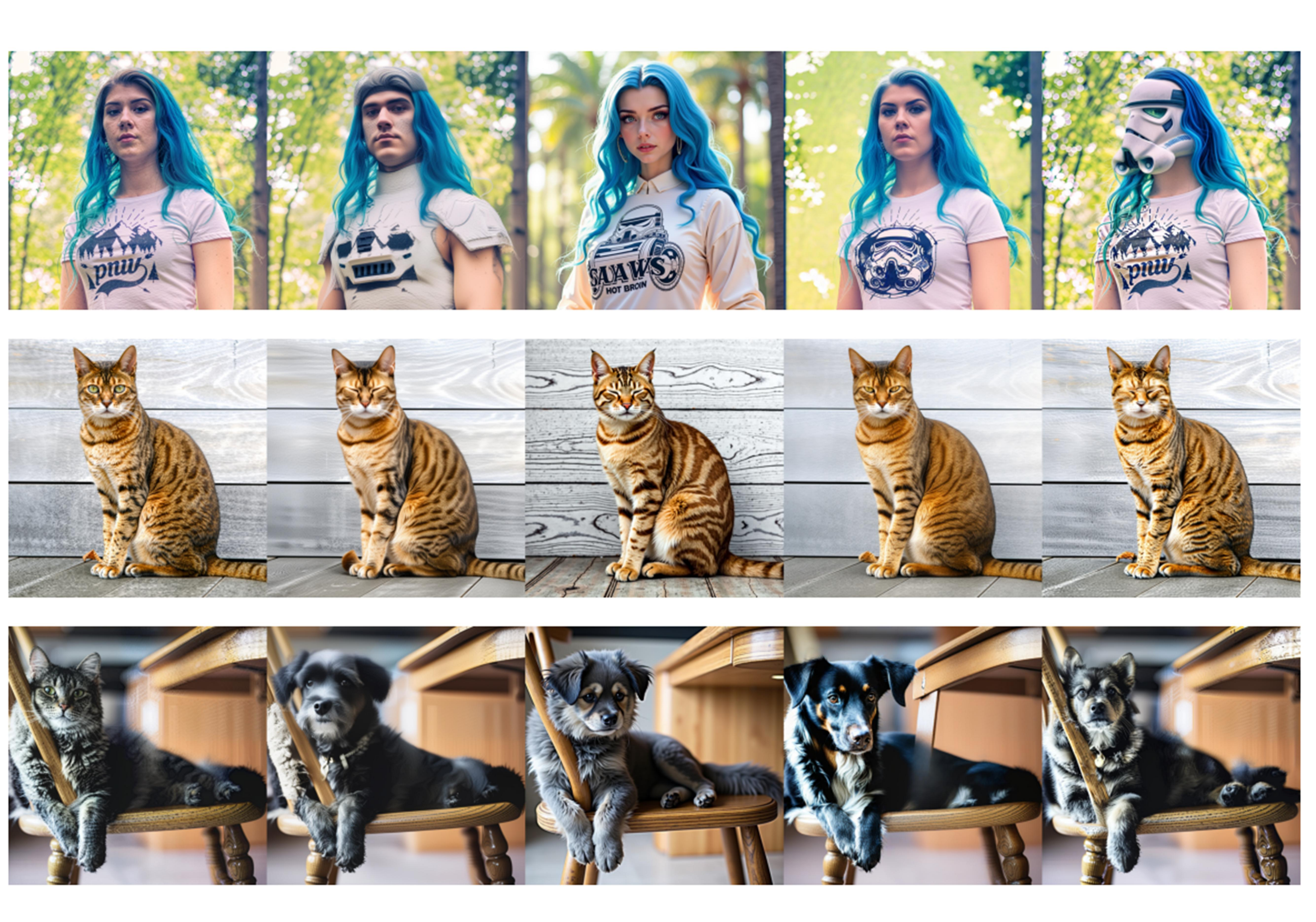}
  };

  \begin{scope}[shift={(image.south west)}, x=1cm, y=1cm]

    \usetikzlibrary{positioning}

    \tikzset{
      colhead/.style={
        font=\fontsize{7pt}{5pt}\editfont,
      },
      rowlbl/.style={
        font=\normalsize,
      }
    }


    \node[rowlbl, anchor=south] (src) at (1.6, 11.2) {Source image $x$};
    \node[rowlbl, right=1.4cm of src] (ledits) {Ledits++~\cite{brack2024ledits++}};
    \node[rowlbl, right=.9cm of ledits, align=center] (turbo)
      {TurboEdit~\cite{deutch2024turboedit}};
    \node[rowlbl, right=.9cm of turbo, align=center] (rfi)
      {RF-Inversion~\cite{routsemantic}};
    \node[rowlbl, right=1.2cm of rfi, align=center] (ours)
      {Ours};

    \node[colhead,  anchor=north,above=0.05cm of src] at (image.south)
      {"A photo of a [\sout{cat} $\rightarrow$\contour{black}{dog}] sitting on a chair."};
    \node[colhead,  anchor=north,above=3.75cm of src] at (image.south)
      {"A photo of a cat with [\sout{open} $\rightarrow$\contour{black}{closed}] eyes"};
    \node[colhead,  anchor=north,above=7.45cm of src] at (image.south)
      {"A photo of a [\sout{woman} $\rightarrow$\contour{black}{stormtrooper}] with blue hair"};
  \end{scope}
\end{tikzpicture}
\caption{\textbf{Qualitative editing results on PIE-1024.} Comparison of Masked Logit-nudging(Ours), Ledits++~\cite{brack2024ledits++}, TurboEdit\cite{deutch2024turboedit} and RF-Inversion~\cite{routsemantic}.}
\label{fig:qualedit1024}
\end{figure*}

Fig.~\ref{fig:qualedit1024} shows additional editing results using the presented MLN approach. 
\subsection{Upscaled PIE-benchmark}
\paragraph{Upsampling Strategy.}
To evaluate edits at 1024px resolution, we require high-quality high-resolution inputs. Since PIE is defined at 512px, we compare two upscaling strategies: \begin{itemize}
\item simple linear interpolation 
\item diffusion-based super-resolution(in the main paper). 
\end{itemize}

The two approaches yield different editing outcomes. Linear interpolation produces overly smooth textures and blurred edges, which propagate into the edited images and lead to less details and can introduce artifacts. In contrast, diffusion-based upsampling reconstructs sharper contours and plausible high-frequency structure, resulting in substantially more faithful and visually coherent edits.

For all our 1024px experiments, we upsample the PIE images using {InvSR}~\cite{yue2025arbitrary} with {4 inference steps}, and additionally provide the {original PIE source prompt} as conditioning to the diffusion-based upsampler. 

\paragraph{Linear Interpolation Baseline.}
For completeness, we also evaluate all editing methods on a naive 1024px variant of the PIE benchmark obtained by {linearly upsampling} the original 512px images.  

As shown in Tab.~\ref{tab:pie_1024_inter}, linearly interpolated inputs in general lead to degraded background preservation and weaker text--image alignment compared to their diffusion-upsampled counterparts.  
These results further highlight that realistic high-frequency reconstruction---as provided by InvSR---is essential for fair and meaningful evaluation of editing performance at 1024px.

\begin{table*}[t]
\centering
\setlength{\tabcolsep}{9pt}
\renewcommand{\arraystretch}{1.10}
\caption{\textbf{Quantitative evaluation (interpolated, i.e. without InvSR) on PIE-1024.}
We report background preservation (PSNR, LPIPS), separate text--image alignment scores (CLIP similarity on whole image and edited region), and wall-clock runtime. 
Best values are \textbf{bold}.}
\label{tab:pie_1024_inter}
\scriptsize
\begin{tabular}{lccccc}
\toprule
\textbf{Method} &
\textbf{PSNR $\uparrow$} &
\textbf{LPIPS $\downarrow$} &
\textbf{CLIP Whole $\uparrow$} &
\textbf{CLIP Edited $\uparrow$} &
\textbf{Wall (s) $\downarrow$} \\
\midrule
PnP~\cite{ju2023direct}                  & 17.67 & 161.96 & 24.62 & 23.20 & 17 \\
LEDITS++~\cite{brack2024ledits++}        & 22.32 & 58.65  & 24.17 & 22.06 & 18.4 \\
ReNoise~\cite{garibi2024renoise}         & 22.33 & 87.33 & 23.32 & 21.11 & 13.2 \\
TurboEdit~\cite{deutch2024turboedit}     & \textbf{28.13} & 43.53 & 25.13 & 22.56 & 4.1 \\
RF-Inversion~\cite{routsemantic}         & 20.62 & 147.41  & 24.41 & 25.11 & 22.3 \\
\textbf{Ours}                            & 27.98 & \textbf{37.76} & \textbf{25.84} & \textbf{23.15} & \textbf{1.6} \\
\bottomrule
\end{tabular}
\end{table*}

\label{sr_benchmark}
\subsection{Recaptioning of OpenImages}
\label{recaptioning}
\label{subsec:ablation_recaption_masks}

\begin{figure*}[t]
    \centering

    \begin{subfigure}[b]{0.32\linewidth}
        \centering
        \includegraphics[width=\linewidth]{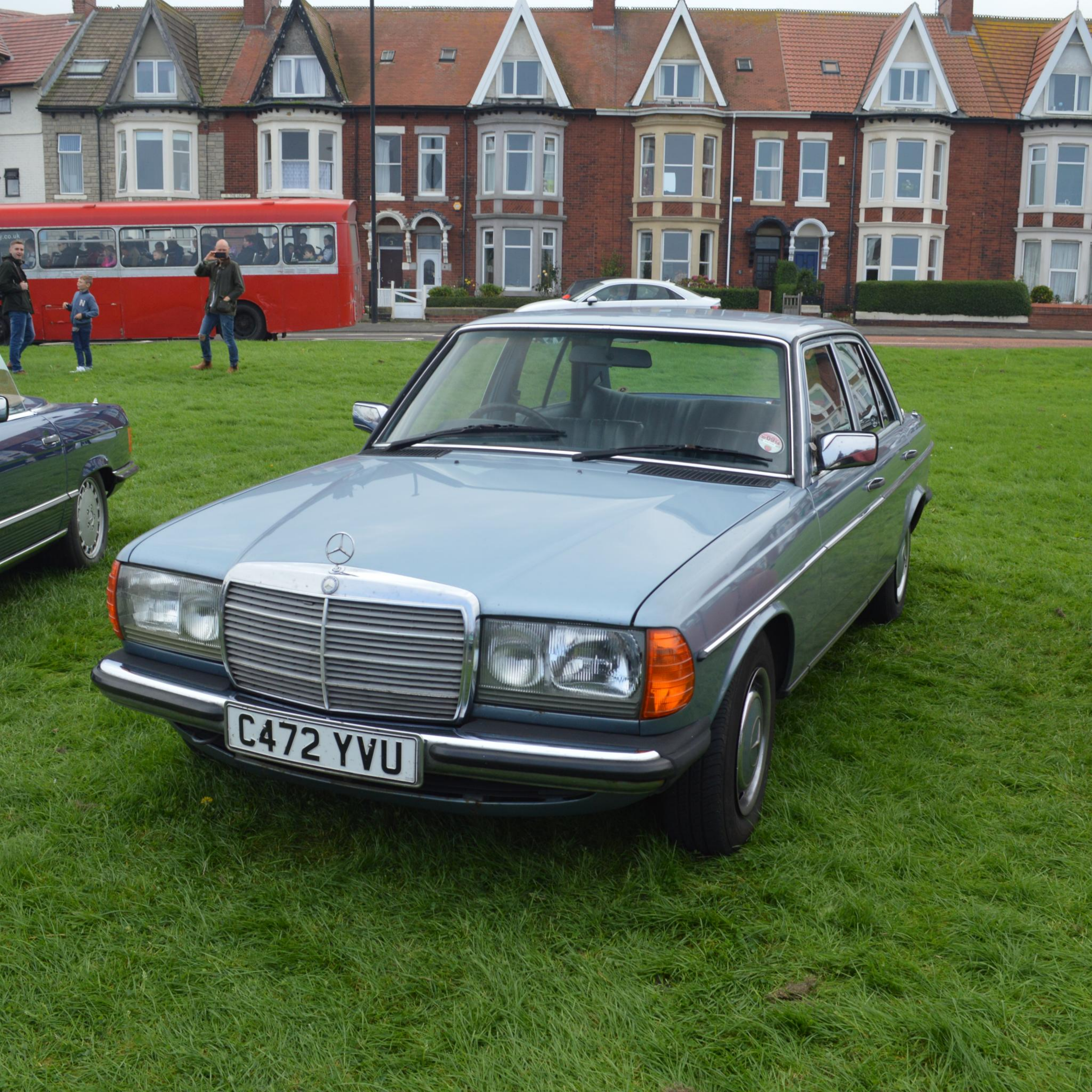}
        \caption*{\small ``A classic light blue sedan parked on a grassy field.''}
    \end{subfigure}
    \hfill
    \begin{subfigure}[b]{0.32\linewidth}
        \centering
        \includegraphics[width=\linewidth]{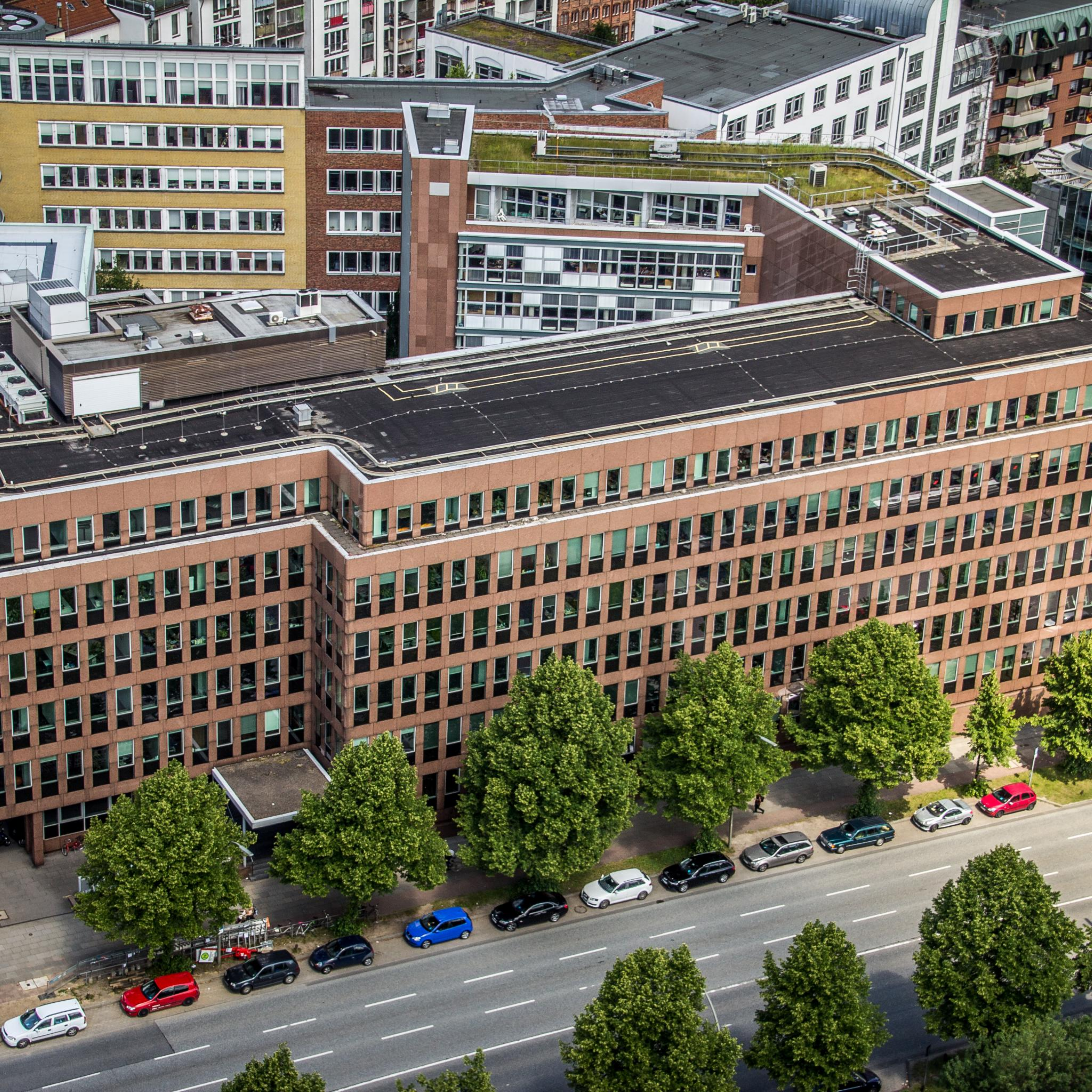}
        \caption*{\small ``An aerial view of a large office building and trees along the street.''}
    \end{subfigure}
    \hfill
    \begin{subfigure}[b]{0.32\linewidth}
        \centering
        \includegraphics[width=\linewidth]{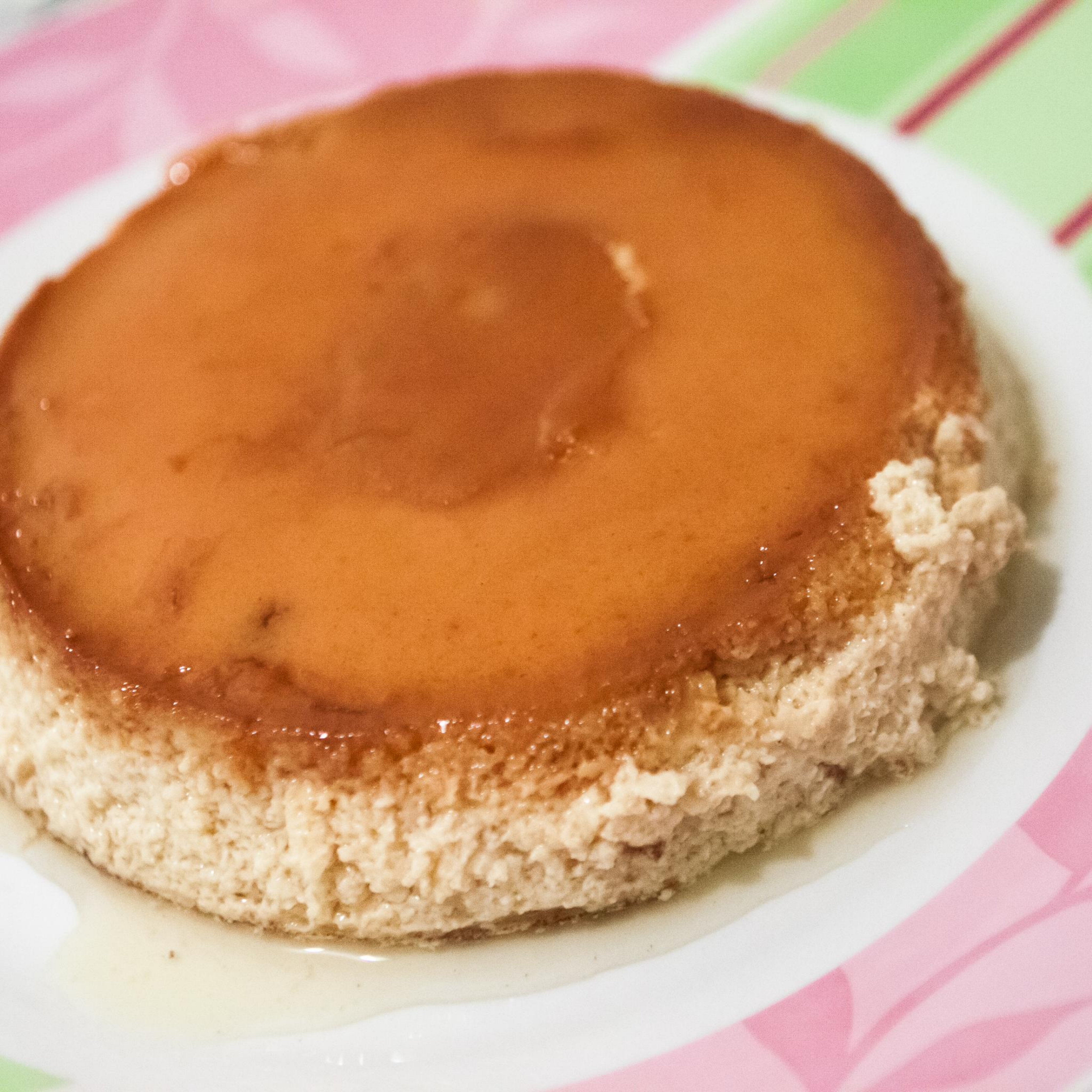}
        \caption*{\small ``A close-up of a caramel flan dessert on a white plate.''}
    \end{subfigure}

    \caption{\textbf{Examples from OpenImages with automatically generated recaptions.}}
    \label{fig:oi_recaption_examples}
\end{figure*}

The OpenImages subset used in our reconstruction evaluations~(\ref{subsec:datasets_reconstruct}) contains no textual annotations.  
To make it usable for text-conditioned training, we automatically generate for each image a language caption using GPT-4V~\cite{zhang2023gpt}.
In Fig.~\ref{fig:oi_recaption_examples} we show 3 recaptioned sample images.

\subsection{Additional qualitative editing samples}
\begin{figure*}[t]
\centering
\begin{tikzpicture}
  \node[inner sep=0] (image) {
    \includegraphics[scale=.156]{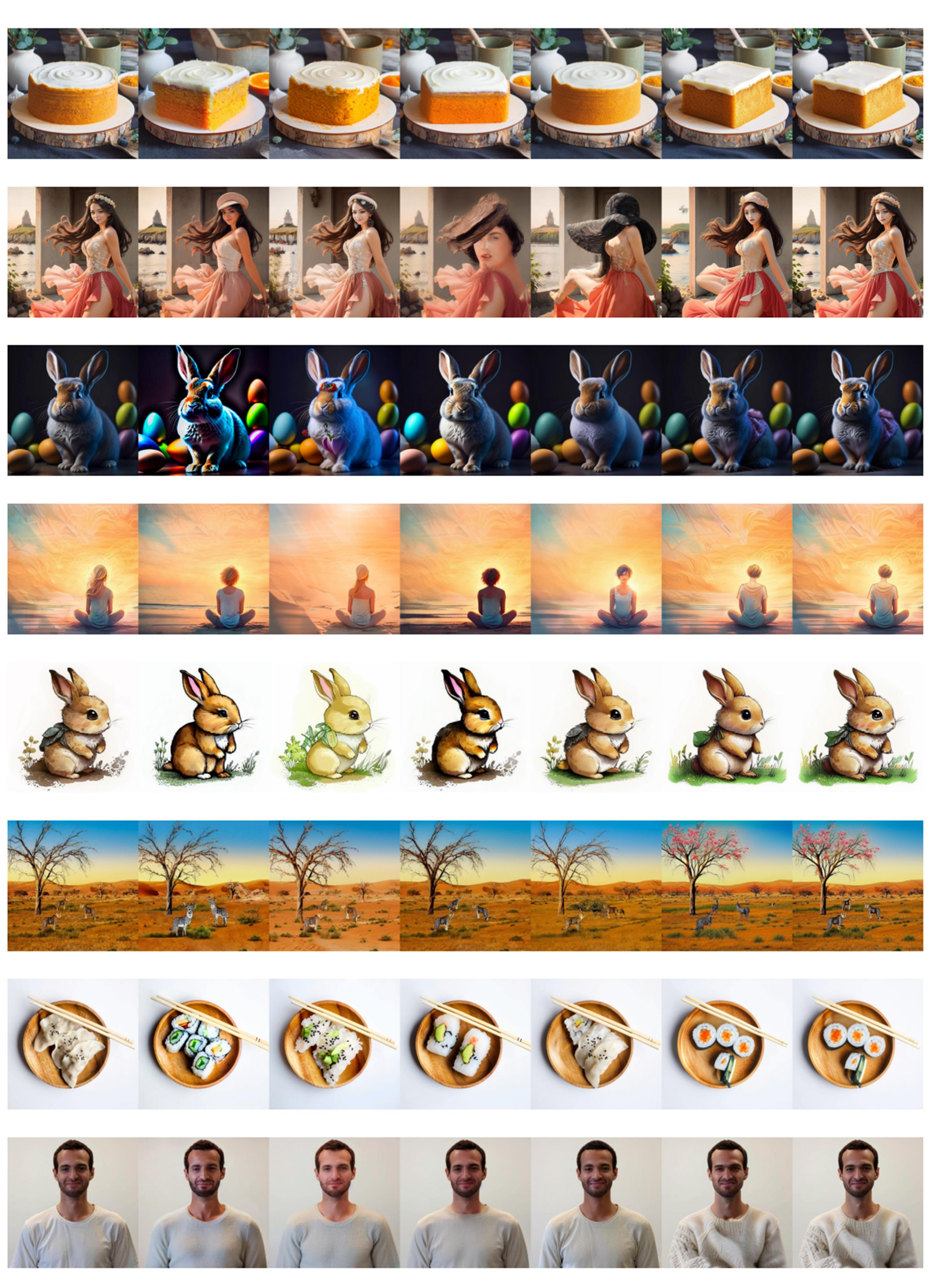}
  };

  \begin{scope}[shift={(image.south west)}, x=1cm, y=1cm]

    \usetikzlibrary{positioning}

    \tikzset{
      colhead/.style={
        font=\fontsize{7pt}{5pt}\editfont,
      },
      rowlbl/.style={
        font=\normalsize,
      }
    }


    \node[rowlbl, anchor=north] (src) at (1.3, 21.1) {Source image $\mathbf{x}$};
    \node[rowlbl, right=.001cm of src] (edf) {EditFriendly~\cite{huberman2024edit}};
    \node[rowlbl, right=.001cm of edf] (pnp) {PnP~\cite{ju2023direct}};
    \node[rowlbl, right=.2cm of pnp] (ledits) {Ledits++~\cite{brack2024ledits++}};
    \node[rowlbl, right=.01cm of ledits, align=center] (turbo)
      {TurboEdit~\cite{deutch2024turboedit}};
    \node[rowlbl, right=.01cm of turbo, align=center] (ours)
      {Ours w/o QR};
\node[rowlbl, right=2cm of turbo, align=center] (ours)
      {Ours w QR};

    \node[colhead,  anchor=north,above=-0.1cm of src] at (image.south)
      {"A man wearing a [\sout{shirt}$\rightarrow$\contour{black}{sweater}]".};
    \node[colhead,  anchor=north,above=2.45cm of src] at (image.south)
      {"three white [\sout{dumplings}$\rightarrow$\contour{black}{sushi}] on brown bowl".};
    \node[colhead,  anchor=north,above=5cm of src] at (image.south)
      {"A group of animals in the desert with a [\sout{dead}$\rightarrow$\contour{black}{blooming}] tree ".};
    \node[colhead,  anchor=north,above=7.65cm of src] at (image.south)
      {"A cartoon drawing of a rabbit on  [\sout{mud}$\rightarrow$\contour{black}{grass}]".};
    \node[colhead,  anchor=north,above=10.15cm of src] at (image.south)
      {"A woman with [\sout{short}$\rightarrow$\contour{black}{long}] hair sitting in the sand at sunset".};
    \node[colhead,  anchor=north,above=12.7cm of src] at (image.south)
      {"A rabbit  [$\rightarrow$\contour{black}{with a dress}] sitting in front of colorful eggs".};
    \node[colhead,  anchor=north,above=15.3cm of src] at (image.south)
      {"A beautiful woman with [\sout{garland}$\rightarrow$\contour{black}{hat}] on head".};
    \node[colhead,  anchor=north,above=17.9cm of src] at (image.south)
      {"A round[\sout{round}$\rightarrow$\contour{black}{square}] cake with orange frosting on a wooden plate".};
  \end{scope}
\end{tikzpicture}
\caption{\textbf{Additional qualitative results on PIE-512.} Editing results of EditFriendly~\cite{huberman2024edit}, PnP~\cite{ju2023direct}, Ledits++~\cite{brack2024ledits++}, TurboEdit~\cite{deutch2024turboedit}, and our proposed Masked Logit Nudging without Quantization refinement (Ours w/o QR) and with Quantization refinement (Ours w QR).}

\label{fig:qualedit512}
\end{figure*}

Additional  qualitative editing results at 512px and 1024px can be seen in Fig.~\ref{fig:qualedit512} and Fig.~\ref{fig:qualedit1024} respectively.
\label{sec:addqualedits1}

\subsection{More ablations}
\label{speed_efficiency}
\label{var_backbones}

\paragraph{Applicability to other VAR backbones}
To evaluate the generality of \textit{Masked Logit Nudging}, we apply it to the {Infinity} model~\cite{han2024infinity} without any retraining or architectural modification. All experiments are conducted at $512\times 512$ resolution with the 2B parameter  checkpoint. Our main  goal is not to optimize Infinity but to verify that MLN transfers across VAR backbones.

Therefore we keep the default infinity hyperparameters identical to those used in the official code and our Switti+MLN implementation. The main difference is that we change $k_{cut}=4$ in the nudging schedule applied.

Due to the difference in quantization schemes—Euclidean residual quantization in SWITTI versus binary spherical quantization (BSQ) in Infinity—we cannot apply the reconstruction enhancement described in Sec.~\ref{section:soft-projection}. Thus, Infinity relies solely on the MLN editing mechanism.

\begin{figure}[t]
\centering
\begin{tikzpicture}
  \node[inner sep=0] (image) {
    \includegraphics[scale=.06]{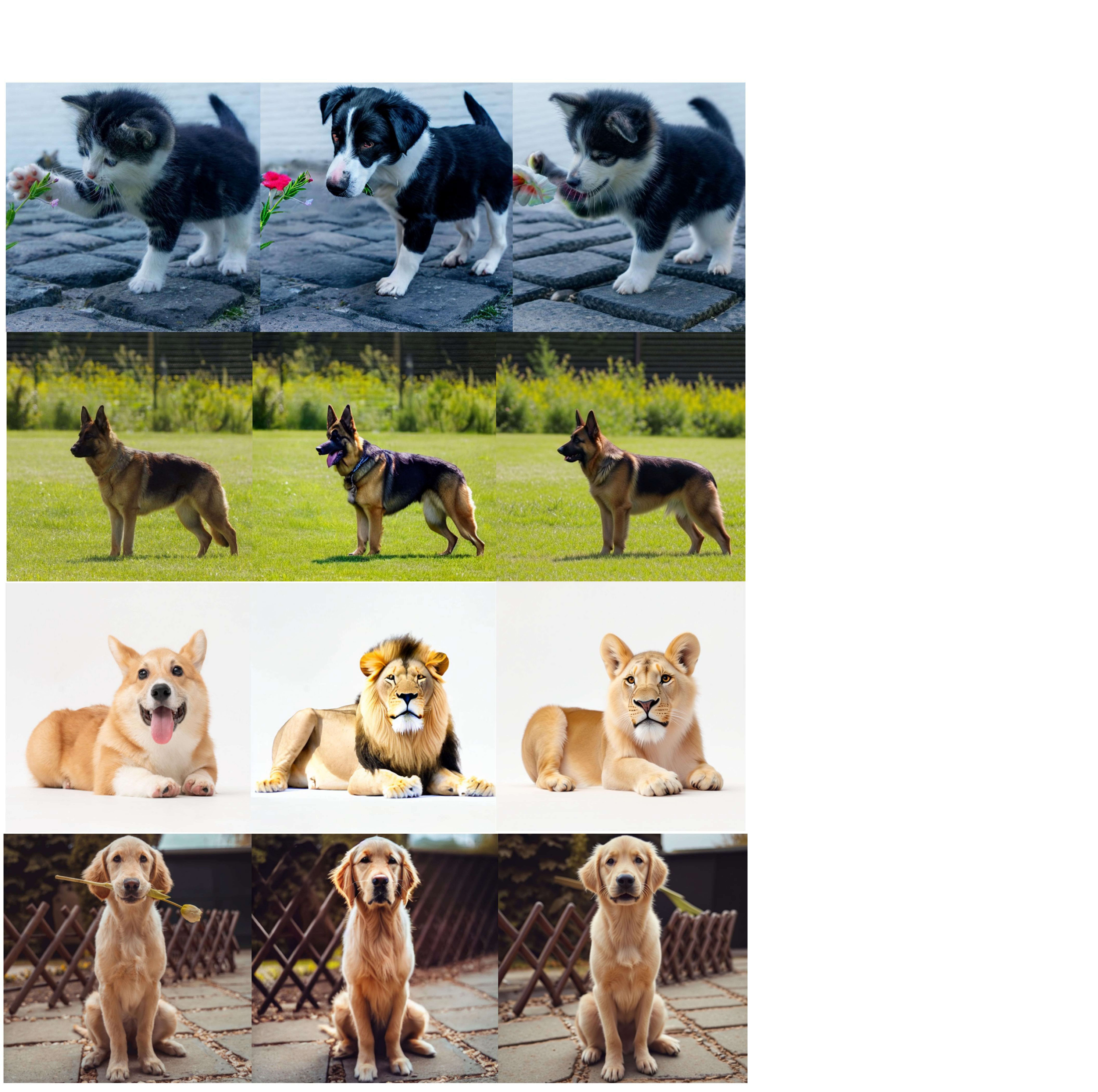}
  };

  \begin{scope}[shift={(image.south west)}, x=1cm, y=1cm]

    \usetikzlibrary{positioning}

    \tikzset{
      colhead/.style={
        font=\fontsize{7pt}{5pt}\editfont,
      },
      rowlbl/.style={
        font=\normalsize,
      }
    }


    \node[rowlbl, anchor=south] (src) at (1.3, 8.2) {Source $\mathbf{x}$};
    \node[rowlbl, right=.1cm of src] (ledits) {Ledits++~\cite{brack2024ledits++}};
    \node[rowlbl, right=-0.2cm of ledits, align=center] (ours)
      {Ours+Infinity~\cite{han2024infinity}};

    \node[colhead,  anchor=north] at (-1,7.2)
      {"[\sout{Dog} $\rightarrow$\contour{black}{cat}]".};
    \node[colhead,  anchor=north,align=center] at (-1,5.1)
      {"[\sout{Closed}$\rightarrow$\contour{black}{Open}]\\ mouth".};
    \node[colhead,  anchor=north] at (-1,3.2)
      {"[\sout{Dog} $\rightarrow$\contour{black}{Lion}]".};
  \node[colhead,  anchor=north,align=center] at (-1,1.2)
      {"[\sout{with Flower}]".};
  \end{scope}
\end{tikzpicture}
\caption{\textbf{Qualitative results on Infinity.} Our approach translates seemlessly to other VAR backbones such as Infinity~\cite{han2024infinity}.}

\label{fig:infinity}
\end{figure}

We provide representative examples for Ledits++~\cite{brack2024ledits++} and Infinity+MLN~(see fig.~\ref{fig:infinity}), demonstrating that MLN reliably transfers across architectures despite quantizer differences.

\paragraph{Precision--Efficiency Trade-Off}
\label{subsec:precision_ablation}

We analyze the impact of numerical precision on runtime and reconstruction fidelity. As shown in Tab.~\ref{tab:ablation_efficiency}, switching from \texttt{float32} to \texttt{float16} substantially accelerates inference---down to only $\sim0.28$\,s per edit---while maintaining competitive reconstruction quality. 

Importantly, \textbf{SWIFTEdit}~\cite{Nguyen_2025_CVPR}, the current state of the art in \emph{fast} image editing, achieves comparable speed but \emph{requires additional training}, whereas our method is entirely \textbf{training-free} and still delivers noticeably better reconstruction fidelity at nearly identical runtime.

Finally, we also evaluate \texttt{float16} at \textbf{1024px} resolution and observe that performance remains stable, confirming that half-precision maintains editability even at high resolutions.

\begin{table}[t]
\centering
\scriptsize
\setlength{\tabcolsep}{3pt}

\caption{\textbf{Precision--efficiency ablation on the PIE-Benchmark.} 
Comparison of editing speed and reconstruction quality for \texttt{float16} and \texttt{float32}. Our method outperforms SWIFTEdit~\cite{Nguyen_2025_CVPR} significantly in overall editing performance, while only requiring 40ms longer and being training-free.}
\label{tab:ablation_efficiency}
\vspace{0.4em}
\begin{tabular}{lccccc}
\toprule
\textbf{Method} & \textbf{Time/Edit (s)} & \textbf{PSNR} $\uparrow$ & \textbf{MSE} $\downarrow$ & \textbf{CLIP} $\uparrow$ \\
\midrule
Ours -- float16 (512px) & 0.28 & 28.01 & 41.23 & 22.01 \\
Ours -- float32 (512px) & 0.82 & 29.70 & 23.30 & 22.72 \\
SWIFTEdit~\cite{Nguyen_2025_CVPR} -- trained, float16 & 0.23 & 23.31 & 61.80 & 21.91 \\
\midrule
Ours -- float16 (1024px) & 0.46 & 24.80 & 74.10 & 22.95 \\ 
{Ours} -- float32 (1024px) & 1.60 & 26.70 & {45.51} & 23.67\\
\bottomrule
\end{tabular}
\end{table}

\subsection{Failure cases}
\label{subsec:mask_failures}

While our method achieves strong editing consistency across most scenarios, we observe that the majority of failure cases arise from {mask inaccuracies}. Since Masked Logit Nudging (MLN) and the Quantization refinement relies on spatial guidance to determine where logits should be nudged toward the source or target distribution, misaligned masks can propagate directly into visible artifacts.

\begin{figure*}[t]
    \centering
    \begin{subfigure}{0.48\linewidth}
        \centering
        \includegraphics[width=\linewidth]{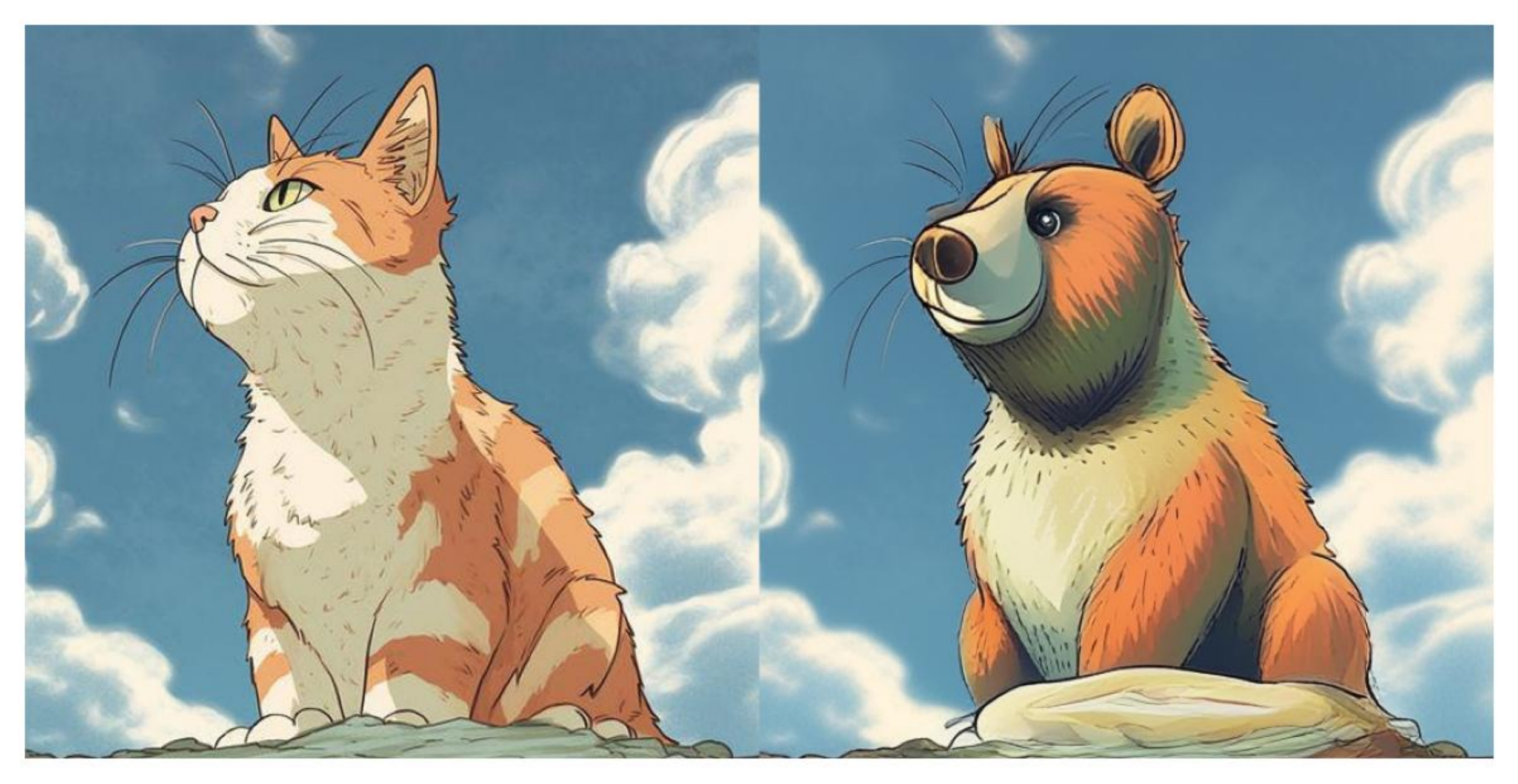}
        \caption*{\textbf{(a) Whisker-level masking error.} 
        When editing a cat into a bear, the mask spills over 
        into the nose hair and whisker regions. As MLN nudges logits inside 
        these pixels, the model unintentionally replaces thin facial details 
        with bear-like textures, producing unnatural local artifacts.}
    \end{subfigure}
    \hfill
    \begin{subfigure}{0.48\linewidth}
        \centering
        \includegraphics[width=\linewidth]{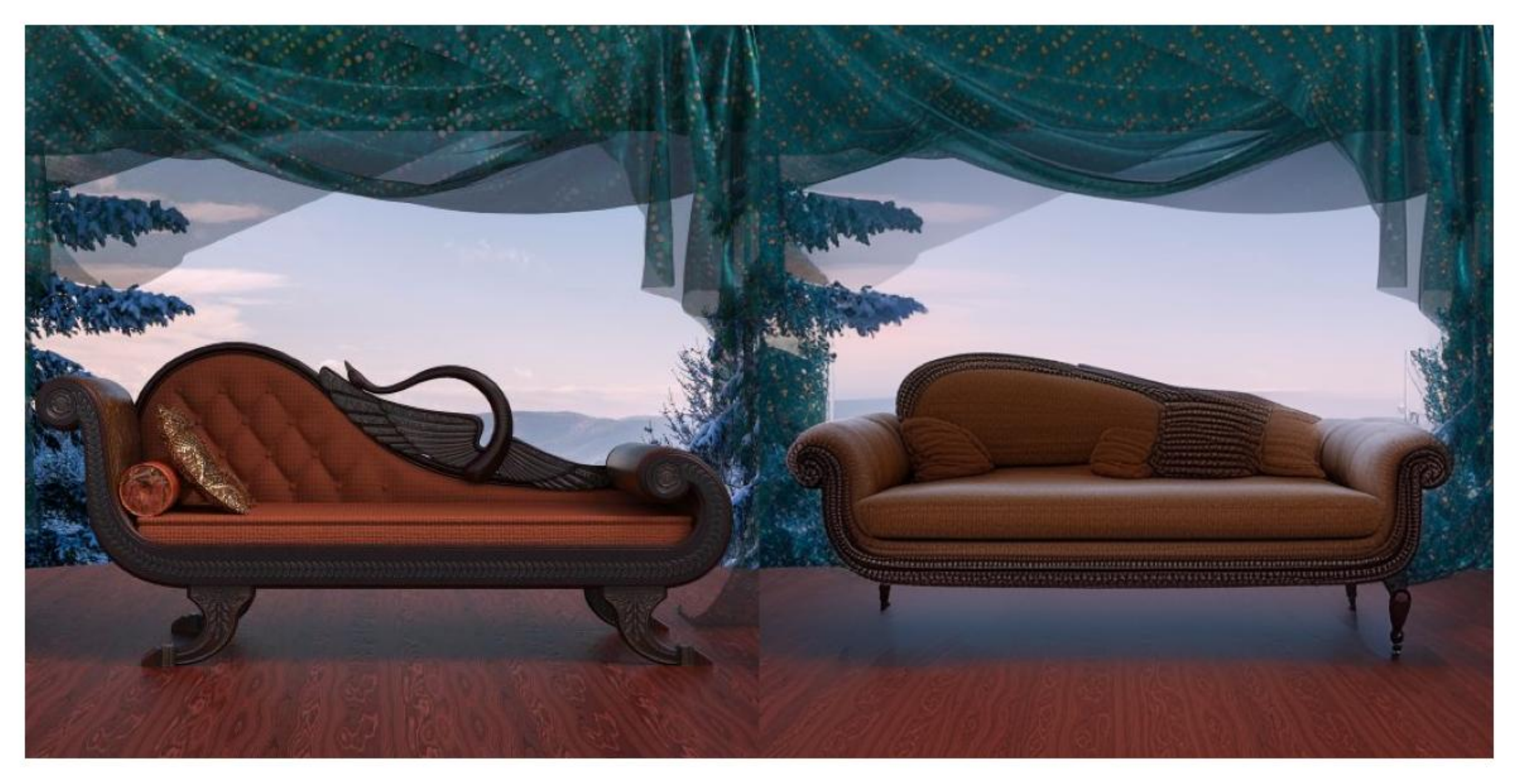}
        \caption*{\textbf{(b) Coarse mask affecting structure.} 
        In this example, the spatial mask extends beyond the edited object 
        and partially covers the sofa boundary. As a result, the AR refinement 
        fails to reconstruct the correct geometry, leading to a warped cushion 
        and inconsistent shading along the couch silhouette.}
    \end{subfigure}

    \vspace{0.5em}
    \caption{\textbf{Masking-related failure cases.}
    Most of our failure modes originate from inaccurate or overly coarse masks. 
    Because MLN modifies logits only inside the predicted editing region, even 
    slight mask misalignments introduce noticeable artifacts—especially in 
    high-frequency areas such as whiskers, fur, or object boundaries. Improving 
    mask precision directly reduces these errors without modifying the underlying 
    editing mechanism.}
    \label{fig:failures}
\end{figure*}

\paragraph{Incorrect fine-grained masks.}
In several challenging examples, the cross-attention–based mask incorrectly assigns high-confidence editing regions to pixels that should remain untouched. Figure~\ref{fig:failures} illustrates such a case: when editing a cat into a bear, the mask partially overlaps with the cat's whiskers and nose hair. As a result, the model unintentionally replaces thin facial details with textures from the target concept, leading to unnatural blending.

\paragraph{Structural errors from coarse masks.}
A second class of failures emerges when the mask captures the correct semantic region but is spatially too coarse. In the couch-editing example, the model attempts to preserve the original geometry, but the spatial mask extends into the background and occludes a portion of the sofa boundary. Consequently, the autoregressive refinement reconstructs a distorted or incomplete couch—either flattening the cushion or introducing inconsistent shading at the edges. These errors confirm that token-level masking must be both semantically accurate and spatially sharp to avoid disrupting the low-frequency structure encoded in the early scales.

\paragraph{Discussion.}
Across both categories, we find that mask quality remains the dominant factor limiting worst-case performance. Since MLN itself operates correctly whenever the preserved region is well specified, improving the mask---e.g., by integrating multi-scale attention cues or leveraging segmentation priors---is likely to further reduce these failure modes without modifying the underlying nudging mechanism.

\label{failure_cases}


\end{document}